%% file: main.tex
\documentclass[letterpaper]{article} 
\usepackage[draft]{aaai25}  
\usepackage{times}  
\usepackage{helvet}  
\usepackage{courier}  
\usepackage[hyphens]{url}  
\usepackage{graphicx} 
\usepackage{natbib}  
\usepackage{caption} 
\usepackage{algorithm}
\usepackage{algorithmic}

\usepackage{newfloat}
\usepackage{listings}
\DeclareCaptionStyle{ruled}{labelfont=normalfont,labelsep=colon,strut=off} 
\floatstyle{ruled}
\newfloat{listing}{tb}{lst}{}
\floatname{listing}{Listing}
\usepackage{color}
\usepackage[dvipsnames]{xcolor}
\usepackage{tikz}
\usetikzlibrary{decorations.text, decorations.pathreplacing}
\usetikzlibrary{calc}
\usetikzlibrary{fit}
\usetikzlibrary{bayesnet}
\usetikzlibrary{arrows}
\usetikzlibrary {arrows.meta}
\usetikzlibrary{positioning}
\usetikzlibrary{shapes}
\usetikzlibrary{shadows}
\usepackage{adjustbox}
\usepackage{amsmath}
\usepackage{amssymb}
\usepackage{booktabs}
\usepackage{multirow}
\usepackage{makecell}
\usepackage{subfigure}

\usepackage{pifont}
\newcommand{\bs}[1]{\boldsymbol{#1}}

\newcommand{\cmark}{\ding{51}}
\newcommand{\xmark}{\ding{55}}
\title{Robust Visual Representation Learning With Multi-modal Prior Knowledge For Image Classification Under Distribution Shift}
\author{
    Hongkuan Zhou\textsuperscript{\rm 1 \rm 2}, Lavdim Halilaj\textsuperscript{\rm 1}, Sebastian Monka\textsuperscript{\rm 1}, Stefan Schmid\textsuperscript{\rm 1}, \\ Yuqicheng Zhu\textsuperscript{\rm 1 \rm 2}, Bo Xiong\textsuperscript{\rm 3}, Steffen Staab\textsuperscript{\rm 2 \rm 4}
}
\affiliations{
    \textsuperscript{\rm 1}Corporate Research, Robert Bosch GmbH, Renningen, Germany\\
    \textsuperscript{\rm 2}University of Stuttgart, Stuttgart, Germany\\
    \textsuperscript{\rm 3}Stanford University, California, US\\
    \textsuperscript{\rm 4}University of Southampton, Southampton, UK\\

    hongkuan.zhou@de.bosch.com
}

\usepackage{bibentry}

\begin{document}

\maketitle
\input{tikz/colors}
\input{tikz/motivation_figure}
\input{sections/00_abstract}
\input{sections/01_introduction}
\input{sections/02_related_works}

\input{sections/03_methodology}

\input{sections/04_experiment}

\input{sections/05_conclusion}

\bibliography{mybibfile}
\input{sections/06_appendix}

\end{document}

%% file: tikz/colors.tex
\definecolor{network-blue}{RGB}{165, 192, 221}
\definecolor{light-yellow}{RGB}{238, 233, 218}
\definecolor{light-green}{RGB}{129, 184, 113}
\definecolor{light-red}{RGB}{242, 182, 160}
\definecolor{light-blue}{RGB}{124, 150, 171}
\definecolor{light-orange}{RGB}{255,147,0}
\definecolor{light-purple}{RGB}{150,115,166}

%% file: tikz/motivation_figure.tex
\tikzstyle{arrow} = [thick,->,>=stealth]
\tikzstyle{myshadow} = [drop shadow={
            shadow scale=0.95,
            shadow xshift=0.5ex,
            shadow yshift=-0.5ex
        }]
\tikzstyle{myshadow2} = [drop shadow={
            shadow scale=0.95,
            shadow xshift=0.3ex,
            shadow yshift=-0.3ex
        }]
\tikzstyle{arrow} = [thick,->,>=stealth]
\begin{figure*}
    \centering
    \begin{adjustbox}{width=0.95\textwidth}
    \begin{tikzpicture}[node distance=2cm]
        \node[inner sep=0pt, label={[label distance=0.0cm, font=\bfseries]90: \small Road Sign Domain}] (distribution-shift-roadsign-train){\includegraphics[width=6cm]{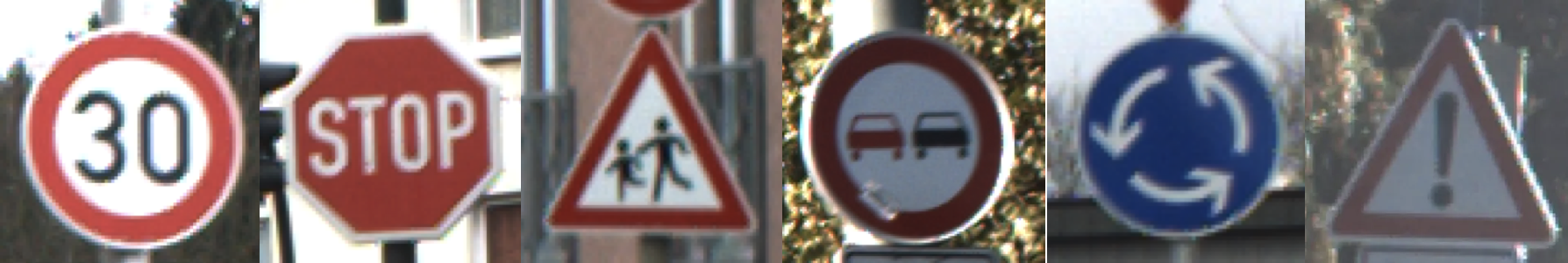}};
        \node[left=1cm of distribution-shift-roadsign-train, yshift=-3.2cm](a){\textbf{(a)}};
        \node[inner sep=0pt, below=0.5cm of distribution-shift-roadsign-train] (distribution-shift-roadsign-test){\includegraphics[width=6cm]{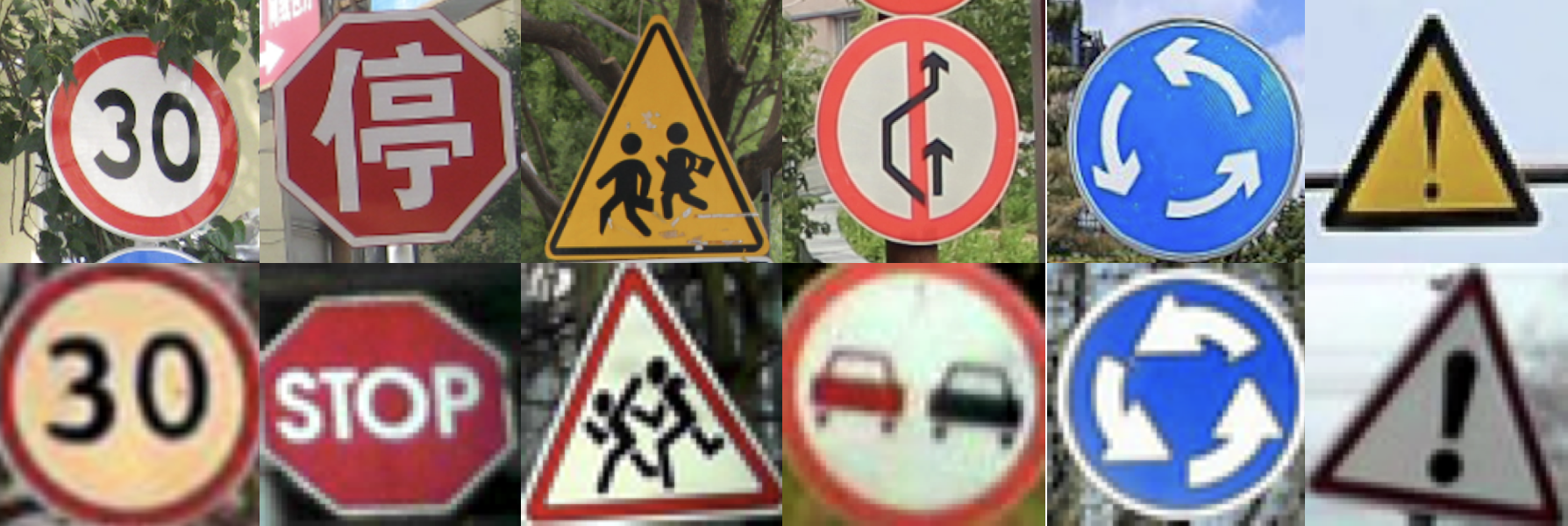}};
        \node[left=0.05cm of distribution-shift-roadsign-train](germany-label){ \small Germany:};
        \node[below=-0.1cm of germany-label](rs_train_label){ \small (Train)};
        \node[left=0.1cm of distribution-shift-roadsign-test, yshift=0.5cm](chinese-label){\small China:};
        \node[below=-0.1cm of chinese-label](rs_test_label){ \small (Test)};
        \node[left=0.05cm of distribution-shift-roadsign-test, yshift=-0.5cm](russia-label){\small Russia:};
        \node[below=-0.1cm of russia-label](rs_test_label){ \small (Test)};
        \draw[dashed, very thick] (-4.5,-3.5) -- (16.5,-3.5);
        \draw[dashed, very thick] (9.75,-3.5) -- (9.75,-8.5);
        \node[inner sep=0pt,  right=1.25cm of distribution-shift-roadsign-train, label={[label distance=0.0cm, font=\bfseries]90: \small Car Domain}] (distribution-shift-dvm-train){\includegraphics[width=5.5cm]{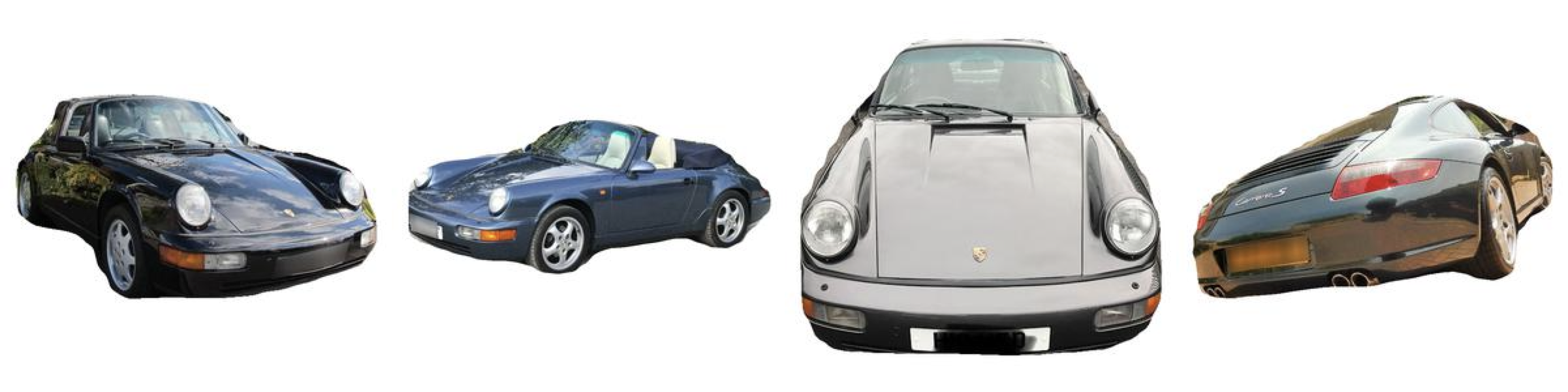}};
        \node[inner sep=0pt,  below=0.5cm of distribution-shift-dvm-train] (distribution-shift-dvm-test){\includegraphics[width=5.5cm]{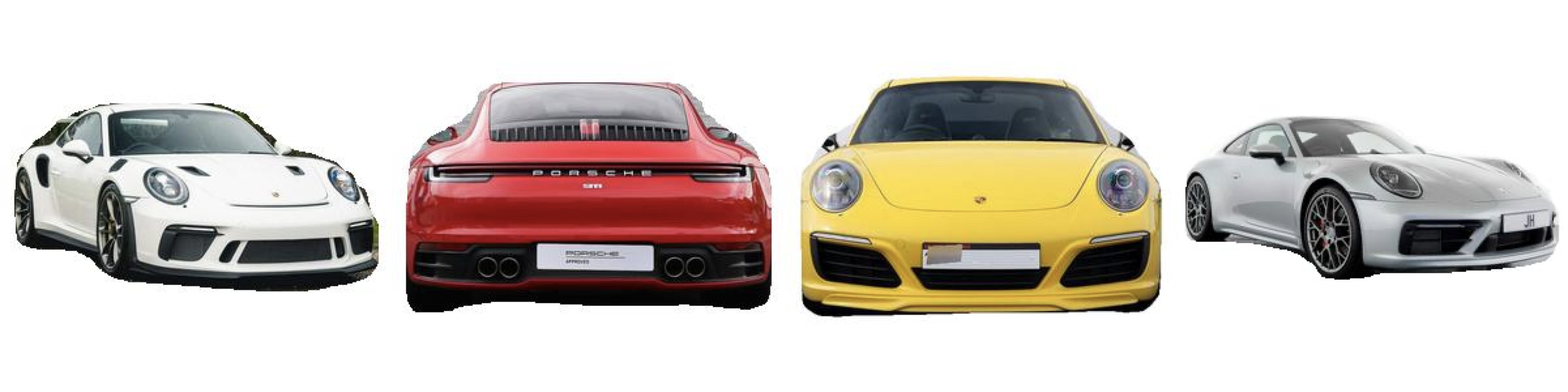}};
        \node[left=0.05cm of distribution-shift-dvm-train](dvm_train){ \small Train:};
        \node[left=0.05cm of distribution-shift-dvm-test](dvm_test){\small Test:};
        \node[inner sep=0pt,  right=2cm of distribution-shift-dvm-train, label={[label distance=0.0cm, font=\bfseries]90: \small ImageNet Domain}] (distribution-shift-imagenet-train){\includegraphics[width=5cm]{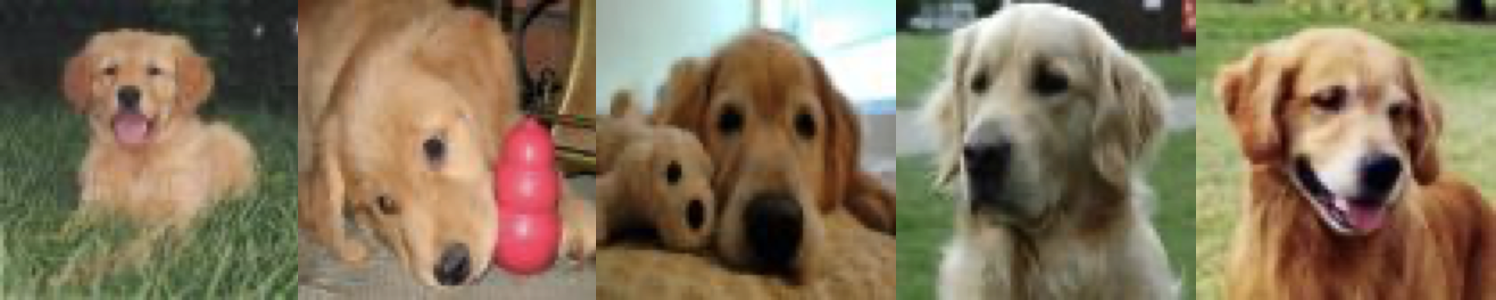}};
        \node[inner sep=0pt,  below=0.5cm of distribution-shift-imagenet-train] (distribution-shift-imagenet-test){\includegraphics[width=5cm]{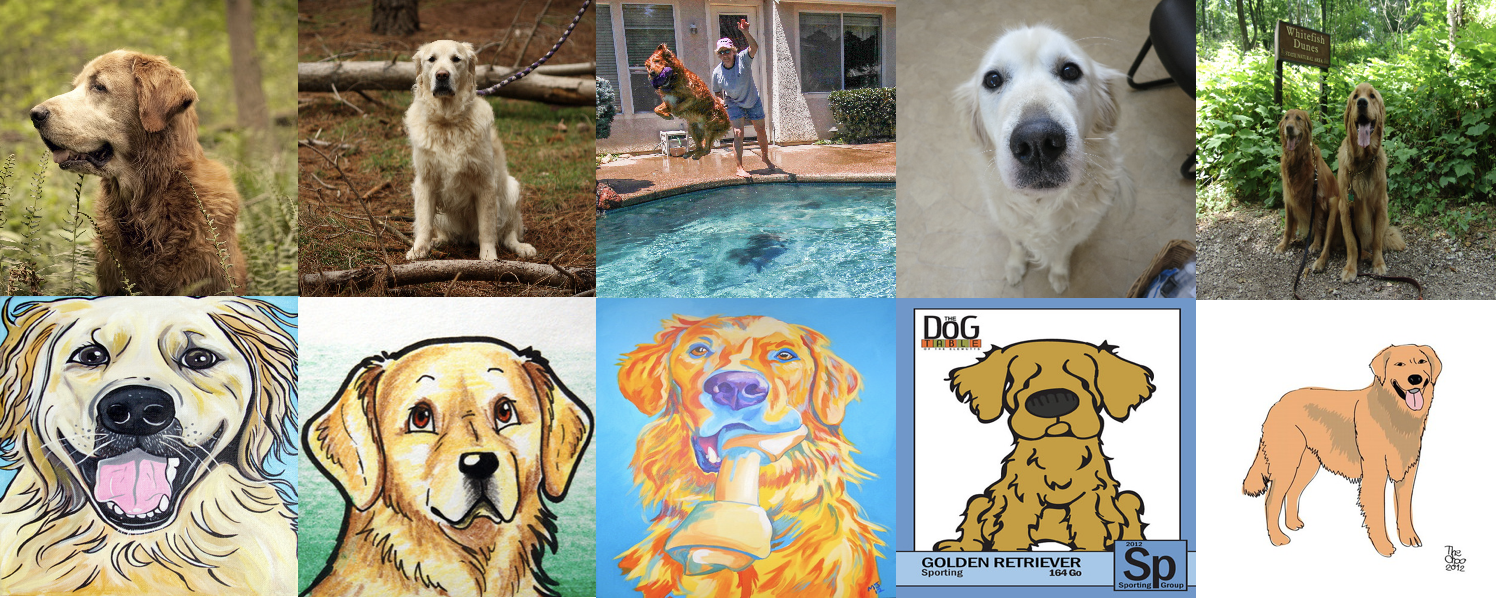}};
        \node[left=0.05cm of distribution-shift-imagenet-train](mini-imagenet){ \small ImageNet:};
        \node[below=-0.1cm of mini-imagenet](train_label){ \small (Train)};
        \node[left=0.05cm of distribution-shift-imagenet-test, yshift=0.5cm](imagenet-v2){\small ImageNet-V2:};
        \node[below=-0.1cm of imagenet-v2](test_label){ \small (Test)};
        \node[left=0.05cm of distribution-shift-imagenet-test, yshift=-0.5cm](imagenet-R){\small ImageNet-R:};
        \node[below=-0.1cm of imagenet-R](test_label){ \small (Test)};  
        \node[inner sep=0pt, below=2.75cm of distribution-shift-roadsign-test, xshift=-1.5cm] (road-sign){\includegraphics[width=1.5cm]{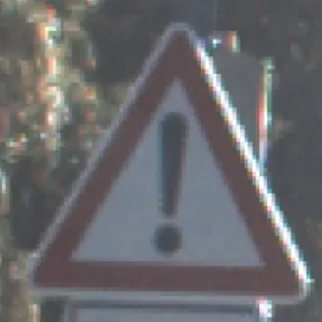}};
        \node[inner sep=0pt, right=1.25cm of road-sign] (road-sign2){\includegraphics[width=1.5cm]{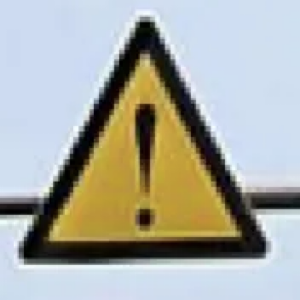}};
        \node[circle, draw=light-orange, thick, above = 0.7cm of road-sign, align=center, minimum size=0.4cm, label={[label distance=0.0cm, font=\bfseries]130: \scriptsize DE}](DE){};
        \draw[dashed, draw=light-purple, thick](road-sign) -- (DE);
        \node[left=1.75cm of road-sign, yshift=2.75cm](b){ \textbf{(b)}};
        \node[circle, draw=light-orange, thick, right = 0.5cm of road-sign, yshift=1cm, align=center, minimum size=0.4cm, label={[label distance=0.0cm, font=\bfseries]90: \scriptsize Triangle}](triangle){};
        \node[circle, draw=light-orange, thick, right = 0.5cm of road-sign, yshift=-1.25cm,  align=center, minimum size=0.4cm, label={[label distance=0.0cm, font=\bfseries]90: \scriptsize Exclamation}](exclamation){};
        \node[circle, draw=light-orange, thick, left = 0.75cm of road-sign, yshift=-1cm, align=center, minimum size=0.4cm, label={[label distance=0.0cm, font=\bfseries]-90: \scriptsize White}](white){};
        \node[circle, draw=light-orange, thick, left = 0.6cm of road-sign, yshift=1cm, align=center, minimum size=0.4cm, label={[label distance=0.0cm, font=\bfseries]90: \scriptsize Red}](red){};
        \draw[draw=light-purple, thick] ([xshift=-0.65cm, yshift=0.5cm]road-sign.east) -- ([xshift=-0.75cm]triangle.west);
        \draw[fill=light-purple!70, draw=light-purple!70] ([xshift=-0.65cm, yshift=0.5cm]road-sign.east) circle (0.04cm);
        \draw[draw=light-purple, arrow] ([xshift=-0.75cm]triangle.west) -- (triangle)node[midway, above, xshift=-0.2cm] {\scriptsize \textcolor{light-purple}{hasShape}};
        \draw[draw=light-purple, thick] ([xshift=-0.75cm, yshift=-0.25cm]road-sign.east) -- ([xshift=-0.8cm]exclamation.west);
        \draw[fill=light-purple!70, draw=light-purple!70] ([xshift=-0.75cm, yshift=-0.25cm]road-sign.east) circle (0.04cm);
        \draw[draw=light-purple, arrow] ([xshift=-0.8cm]exclamation.west) -- (exclamation)node[midway, below, xshift=-0.4cm, yshift=0.05cm] {\scriptsize \textcolor{light-purple}{hasSignLegend}};
        \draw[draw=light-purple, thick] ([xshift=0.6cm, yshift=-0.25cm]road-sign.west) -- ([xshift=1cm]white.east);
        \draw[fill=light-purple!70, draw=light-purple!70] ([xshift=0.6cm, yshift=-0.25cm]road-sign.west) circle (0.04cm);
        \draw[draw=light-purple, arrow] ([xshift=1cm]white.east) -- (white)node[midway, below, xshift=0.4cm, yshift=0.05cm] {\scriptsize \textcolor{light-purple}{hasBackgroundC.}};
        \draw[draw=light-purple, thick] ([xshift=0.6cm, yshift=0.25cm]road-sign.west) -- ([xshift=1cm]red.east);
        \draw[fill=light-purple!70, draw=light-purple!70] ([xshift=0.6cm, yshift=0.25cm]road-sign.west) circle (0.04cm);
        \draw[draw=light-purple, arrow] ([xshift=1cm]red.east) -- (red)node[midway, above, xshift=0.3cm, yshift=-0.05cm] {\scriptsize \textcolor{light-purple}{hasBorderC.}};
        \node (shape-triangle-image-box) [rectangle, draw=black, below = 0.1cm of triangle, minimum width = 1.5cm, minimum height=0.6cm, fill=white]{};
        \node[inner sep=0pt, left of=shape-triangle-image-box, myshadow2, xshift=1.55cm, yshift=0cm] (shape-triangle-1){\includegraphics[width=0.4cm, height=0.4cm]{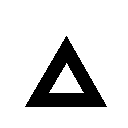}};
        \node[inner sep=0pt, right = 0.1cm of shape-triangle-1, myshadow2, label={[label distance=0.0cm]0: \small ...}] (shape-triangle-2){\includegraphics[width=0.4cm, height=0.4cm]{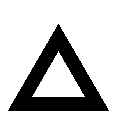}};
        \node (color-red-image-box) [rectangle, draw=black, below = 0.1cm of red, minimum width = 1.5cm, minimum height=0.6cm]{};
        \node[inner sep=0pt, left of=color-red-image-box, myshadow2, xshift=1.55cm, yshift=0cm] (color-red-1){\includegraphics[width=0.4cm, height=0.4cm]{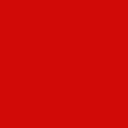}};
        \node[inner sep=0pt, right = 0.1cm of color-red-1, myshadow2, label={[label distance=0.0cm]0: \small ...}] (color-red-2){\includegraphics[width=0.4cm, height=0.4cm]{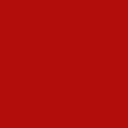}};
        \node (legend-exclamation-image-box) [rectangle, draw=black, below = 0.1cm of exclamation, minimum width = 1.5cm, minimum height=0.6cm]{};
        \node[inner sep=0pt, left of=legend-exclamation-image-box, myshadow2, xshift=1.55cm, yshift=0cm] (legend-exclamation-1){\includegraphics[width=0.4cm, height=0.4cm]{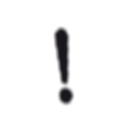}};
        \node[inner sep=0pt, right = 0.1cm of legend-exclamation-1, myshadow2, label={[label distance=0.0cm]0: \small ...}] (legend-exclamation-2){\includegraphics[width=0.4cm, height=0.4cm]{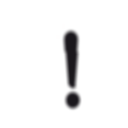}};
        \node (color-white-image-box) [rectangle, draw=black, above = 0.1cm of white, minimum width = 1.5cm, minimum height=0.6cm]{};
        \node[inner sep=0pt, left of=color-white-image-box, myshadow2, xshift=1.55cm, yshift=0cm] (color-white-1){\includegraphics[width=0.4cm, height=0.4cm]{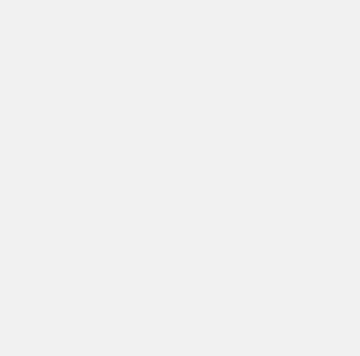}};
        \node[inner sep=0pt, right = 0.1cm of color-white-1, myshadow2, label={[label distance=0.0cm]0: \small ...}] (color-white-2){\includegraphics[width=0.4cm, height=0.4cm]{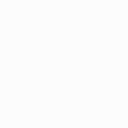}};
        \draw [draw=light-purple, thick] (legend-exclamation-image-box) -- (exclamation);
        \draw [draw=light-purple, thick] (color-red-image-box) -- (red);
        \draw [draw=light-purple, thick] (color-white-image-box) -- (white);
        \draw [draw=light-purple, thick] (shape-triangle-image-box) -- (triangle);

        \node[circle, draw=light-orange, thick, above = 0.7cm of road-sign2, align=center, minimum size=0.4cm, label={[label distance=0.0cm, font=\bfseries]50: \scriptsize CN}](CN){};
        \node[circle, draw=light-orange, thick, above = 0.2cm of CN, xshift=-1.3cm, align=center, minimum size=0.4cm, label={[label distance=0.0cm, font=\bfseries]90: \scriptsize Danger Sign}](danger-sign){};
        \draw[dashed, draw=light-purple, thick](road-sign2) -- (CN);
        \draw[arrow, dashed, draw=light-purple, thick](CN) -- (danger-sign);
        \draw[arrow, dashed, draw=light-purple, thick](DE) -- (danger-sign);
        \draw[draw=light-purple, thick] ([xshift=0.65cm, yshift=0.5cm]road-sign2.west) -- ([xshift=0.75cm]triangle.east);
        \draw[fill=light-purple!70, draw=light-purple!70] ([xshift=0.65cm, yshift=0.5cm]road-sign2.west) circle (0.04cm);
        \draw[draw=light-purple, arrow] ([xshift=0.75cm]triangle.east) -- (triangle)node[midway, above, xshift=0.2cm] {\scriptsize \textcolor{light-purple}{hasShape}};
        \draw[draw=light-purple, thick] ([xshift=-0.75cm, yshift=-0.1cm]road-sign2.east) -- ([xshift=0.6cm]exclamation.east);
        \draw[fill=light-purple!70, draw=light-purple!70] ([xshift=-0.75cm, yshift=-0.1cm]road-sign2.east) circle (0.04cm);
        \draw[draw=light-purple, arrow] ([xshift=0.6cm]exclamation.east) -- (exclamation)node[midway, below, xshift=0.4cm, yshift=0.05cm] {\scriptsize \textcolor{light-purple}{hasSignLegend}};
        \node[circle, draw=light-orange, thick, right = 0.5cm of road-sign2, yshift=1cm, align=center, minimum size=0.4cm, label={[label distance=0.0cm, font=\bfseries]90: \scriptsize Black}](black){};
        \node[circle, draw=light-orange, thick, right = 0.75cm of road-sign2, yshift=-1cm, align=center, minimum size=0.4cm, label={[label distance=0.0cm, font=\bfseries]-90: \scriptsize Yellow}](yellow){};
        \node (color-yellow-image-box) [rectangle, draw=black, above = 0.1cm of yellow, minimum width = 1.5cm, minimum height=0.6cm]{};
        \node[inner sep=0pt, left of=color-yellow-image-box, myshadow2, xshift=1.55cm, yshift=0cm] (color-yellow-1){\includegraphics[width=0.4cm, height=0.4cm]{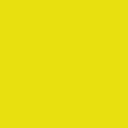}};
        \node[inner sep=0pt, right = 0.1cm of color-yellow-1, myshadow2, label={[label distance=0.0cm]0: \small ...}] (color-yellow-2){\includegraphics[width=0.4cm, height=0.4cm]{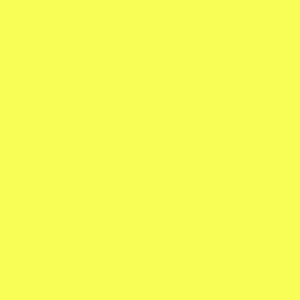}};
        \node (color-black-image-box) [rectangle, draw=black, below = 0.1cm of black, minimum width = 1.5cm, minimum height=0.6cm, fill=white]{};
        \draw[draw=light-purple, thick] ([xshift=-0.5cm, yshift=-0.25cm]road-sign2.east) -- ([xshift=-1cm]yellow.west);
        \draw[fill=light-purple, draw=light-purple] ([xshift=-0.5cm, yshift=-0.25cm]road-sign2.east) circle (0.04cm);
        \draw[draw=light-purple, arrow] ([xshift=-1cm]yellow.west) -- (yellow)node[midway, below, xshift=-0.4cm, yshift=0.05cm] {\scriptsize \textcolor{light-purple}{hasBackgroundC.}};
        \node[inner sep=0pt, left of=color-black-image-box, myshadow2, xshift=1.55cm, yshift=0cm] (color-black-1){\includegraphics[width=0.4cm, height=0.4cm]{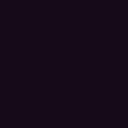}};
        \node[inner sep=0pt, right = 0.1cm of color-black-1, myshadow2, label={[label distance=0.0cm]0: \small ...}] (color-black-2){\includegraphics[width=0.4cm, height=0.4cm]{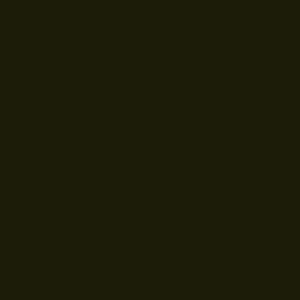}};
        \draw[draw=light-purple, thick] ([xshift=-0.6cm, yshift=0.25cm]road-sign2.east) -- ([xshift=-1cm]black.west);
        \draw[fill=light-purple!70, draw=light-purple!70] ([xshift=-0.6cm, yshift=0.25cm]road-sign2.east) circle (0.04cm);
        \draw[draw=light-purple, arrow] ([xshift=-1cm]black.west) -- (black)node[midway, above, xshift=-0.1cm, yshift=-0.05cm] {\scriptsize \textcolor{light-purple}{hasBorderC.}};
        \draw [draw=light-purple, thick] (color-black-image-box) -- (black);
        \draw [draw=light-purple, thick] (color-yellow-image-box) -- (yellow);
        \node[inner sep=0pt, right=4cm of road-sign2, yshift=-0.5cm] (porsche){\includegraphics[width=1.5cm]{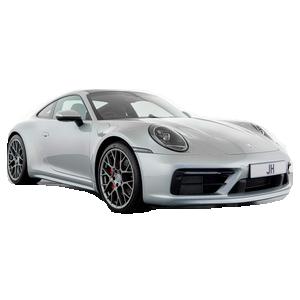}};
        \node[inner sep=0pt, right=4cm of road-sign2, yshift=1.5cm] (porsche2){\includegraphics[width=1.5cm]{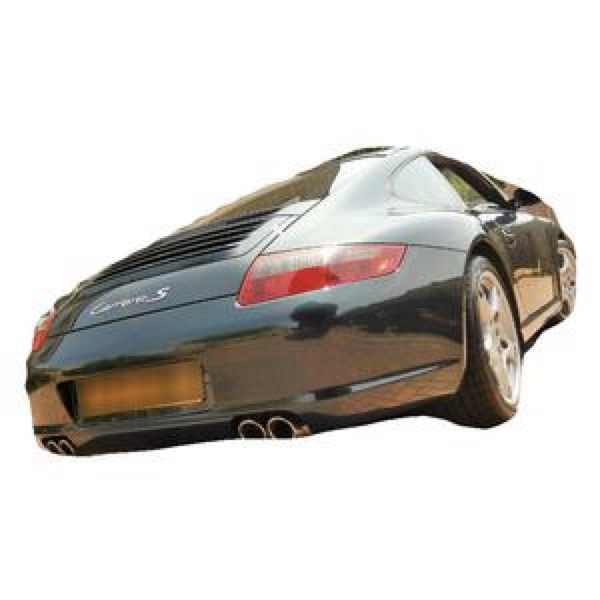}};
        \node (silver) [circle, minimum size=0.4cm, draw=light-orange, thick, right = 0.5cm of porsche, yshift=-1.25cm, label={[label distance=0cm, font=\bfseries]-90: \scriptsize Silver}]{};
        \draw[draw=light-purple, thick] ([xshift=-0.5cm, yshift=0cm]porsche.east) -- ([xshift=-0.75cm]silver.west);
        \draw[fill=light-purple!70, draw=light-purple!70] ([xshift=-0.5cm, yshift=0cm]porsche.east) circle (0.04cm);
        \draw[draw=light-purple, arrow] ([xshift=-0.75cm]silver.west) -- (silver)node[midway, below, xshift=-0.2cm] {\scriptsize \textcolor{light-purple}{hasColor}};
        \node (color-silver-image-box) [rectangle, draw=black, above = 0.1cm of silver, minimum width = 1.5cm, minimum height=0.6cm]{};
        \node[inner sep=0pt, left of=color-silver-image-box, xshift=1.55cm, yshift=0cm] (color-silver-1){\includegraphics[width=0.4cm, height=0.4cm]{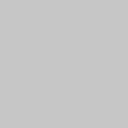}};
        \node[inner sep=0pt, right = 0.1cm of color-silver-1, label={[label distance=0.0cm]0: \small ...}] (color-silver-2){\includegraphics[width=0.4cm, height=0.4cm]{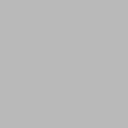}};
        \draw[draw=light-purple, thick] (silver) -- (color-silver-image-box);
        \node (coupe) [circle, minimum size=0.4cm, draw=light-orange, thick, left = 0.5cm of porsche, yshift=0.5cm, label={[label distance=0cm, font=\bfseries, xshift=-0.2cm]90: \scriptsize Coupe}]{};
        \node (bodytype-coupe-image-box) [rectangle, draw=black, below = 0.1cm of coupe, minimum width = 1.5cm, minimum height=0.6cm]{};
        \node[inner sep=0pt, left of=bodytype-coupe-image-box, myshadow2, xshift=1.55cm, yshift=0cm] (bodytype-coupe-1){\includegraphics[width=0.4cm, height=0.4cm]{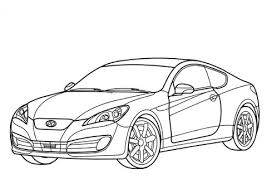}};
        \node[inner sep=0pt, right = 0.1cm of bodytype-coupe-1, myshadow2, label={[label distance=0.0cm]0: \small ...}] (bodytype-coupe-2){\includegraphics[width=0.4cm, height=0.4cm]{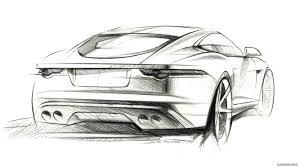}};
        \draw[draw=light-purple, thick] (coupe) -- (bodytype-coupe-image-box);
        \draw[draw=light-purple, thick] ([xshift=0.5cm, yshift=0.25cm]porsche.west) -- ([xshift=0.75cm]coupe.east);
        \draw[fill=light-purple!70, draw=light-purple!70] ([xshift=0.5cm, yshift=0.25cm]porsche.west) circle (0.04cm);
        \draw[draw=light-purple, arrow] ([xshift=0.75cm]coupe.east) -- (coupe)node[midway, above, xshift=0.4cm, yshift=-0.05cm] {\scriptsize \textcolor{light-purple}{hasBodytype}};

        \draw[draw=light-purple, thick] ([xshift=0.5cm]porsche2.west) -- ([xshift=-0.2cm]porsche2.west);
        \draw[fill=light-purple!70, draw=light-purple!70] ([xshift=0.5cm]porsche2.west) circle (0.04cm);
        \draw[draw=light-purple, thick, arrow]([xshift=-0.2cm]porsche2.west) -- (coupe.north east)node[midway, above, rotate=73, xshift=0.2cm] {\scriptsize \textcolor{light-purple}{hasBodytype}};
        
        \node (auto) [circle, minimum size=0.4cm, draw=light-orange, thick, left = 0.75cm of porsche, yshift=-1cm, label={[label distance=0cm, font=\bfseries]-90: \scriptsize Auto}]{};
        \draw[draw=light-purple, thick] ([xshift=0.5cm, yshift=-0.1cm]porsche.west) -- ([xshift=1cm]auto.east);
        \draw[fill=light-purple!70, draw=light-purple!70] ([xshift=0.5cm, yshift=-0.1cm]porsche.west) circle (0.04cm);
        \draw[draw=light-purple, arrow] ([xshift=1cm]auto.east) -- (auto)node[midway, below, xshift=0.4cm, yshift=-0.05cm] {\scriptsize \textcolor{light-purple}{hasGearType}};
        \node (viewpoint45) [circle, minimum size=0.4cm, draw=light-orange, thick, right = 1cm of porsche, yshift=0.2cm, label={[label distance=-0.1cm, font=\bfseries]-40: \scriptsize 45$^{\circ}$}]{};
        \draw[draw=light-purple, thick] ([xshift=-0.3cm, yshift=-0.1cm]porsche.east) -- ([xshift=-1cm]viewpoint45.east);
        \draw[fill=light-purple!70, draw=light-purple!70] ([xshift=-0.3cm, yshift=-0.1cm]porsche.east) circle (0.04cm);
        \draw[draw=light-purple, arrow] ([xshift=-1cm]viewpoint45.east) -- (viewpoint45)node[midway, above, xshift=-0.4cm, yshift=-0.05cm] {\scriptsize \textcolor{light-purple}{hasViewpoint}};
        \node (number2) [circle, minimum size=0.4cm, draw=light-orange, thick, right = 0.5cm of porsche, yshift=1cm, label={[label distance=0cm, font=\bfseries]-90: \scriptsize Number2}]{};
        \draw[draw=light-purple, thick] ([xshift=-0.7cm, yshift=0.1cm]porsche.east) -- ([xshift=-1.25cm]number2.east);
        \draw[fill=light-purple!70, draw=light-purple!70] ([xshift=-0.7cm, yshift=0.1cm]porsche.east) circle (0.04cm);
        \draw[draw=light-purple, arrow] ([xshift=-1.25cm]number2.east) -- (number2)node[midway, above, xshift=-0.4cm, yshift=-0.05cm] {\scriptsize \textcolor{light-purple}{hasDoorNumber}};
        \draw[draw=light-purple, thick] ([xshift=-0.5cm]porsche2.east) -- ([xshift=0.5cm]porsche2.east);
        \draw[fill=light-purple!70, draw=light-purple!70] ([xshift=-0.5cm]porsche2.east) circle (0.04cm);
        \draw[draw=light-purple, thick, arrow]([xshift=0.5cm]porsche2.east) -- (number2.north)node[midway, above, rotate=-73, xshift=-0.2cm] {\scriptsize \textcolor{light-purple}{hasDoorN.}};

        \node (viewpoint135) [circle, minimum size=0.4cm, draw=light-orange, thick, right = 1cm of porsche2, yshift=0.7cm, label={[label distance=-0.1cm, font=\bfseries]-40: \scriptsize 135$^{\circ}$}]{};
        \draw[draw=light-purple, thick] ([xshift=-0.4cm, yshift=0.1cm]porsche2.east) -- ([xshift=-1cm]viewpoint135.east);
        \draw[fill=light-purple!70, draw=light-purple!70] ([xshift=-0.4cm, yshift=0.1cm]porsche2.east) circle (0.04cm);
        \draw[draw=light-purple, arrow] ([xshift=-1cm]viewpoint135.east) -- (viewpoint135)node[midway, above, xshift=-0.4cm, yshift=-0.05cm] {\scriptsize \textcolor{light-purple}{hasViewpoint}};

        \node (manuell) [circle, minimum size=0.4cm, draw=light-orange, thick, left = 0.75cm of porsche2, yshift=0.75cm, label={[label distance=0cm, font=\bfseries]-90: \scriptsize Manuell}]{};
        \draw[draw=light-purple, thick] ([xshift=0.5cm, yshift=0.2cm]porsche2.west) -- ([xshift=1cm]manuell.east);
        \draw[fill=light-purple!70, draw=light-purple!70] ([xshift=0.5cm, yshift=0.2cm]porsche2.west) circle (0.04cm);
        \draw[draw=light-purple, arrow] ([xshift=1cm]manuell.east) -- (manuell)node[midway, above, xshift=0.4cm, yshift=-0.05cm] {\scriptsize \textcolor{light-purple}{hasGearType}};
        \node (i_node1) [circle, minimum size=0.4cm, draw=light-orange, thick, right = 6cm of porsche, yshift=2.7cm, label={[label distance=0cm, font=\bfseries]90: \scriptsize Things}]{};
        \node[left=3cm of i_node1, yshift=0.5cm](c){\textbf{(c)}};
        \node (i_node1_1) [circle, minimum size=0.4cm, draw=light-orange, thick, below = 0.3cm of i_node1, xshift=-2cm, label={[label distance=0cm, font=\bfseries, xshift=0cm]90: \scriptsize Food}]{};
        \node (i_node1_1_1) [circle, minimum size=0.4cm, draw=light-orange, thick, below = 0.3cm of i_node1_1, xshift=-0.5cm, label={[label distance=-0.1cm, font=\bfseries, xshift=0cm]110: \scriptsize Beverage}]{};
        \node (i_node1_1_1_1) [circle, minimum size=0.4cm, draw=light-orange, thick, below = 0.3cm of i_node1_1_1, xshift=-0.4cm, label={[label distance=0cm, font=\bfseries, xshift=0cm]180: \scriptsize Beer}]{};
        \node (beer-image-box) [rectangle, draw=black, below = 0.5cm of i_node1_1_1_1, minimum width = 1.5cm, minimum height=0.6cm]{};
        \node[inner sep=0pt, left of=beer-image-box, myshadow2, xshift=1.55cm, yshift=0cm] (beer-1){\includegraphics[width=0.4cm, height=0.4cm]{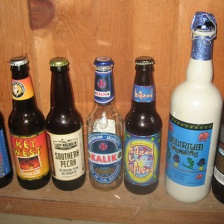}};
        \node[inner sep=0pt, right = 0.1cm of beer-1, myshadow2, label={[label distance=0.0cm]0: \small ...}] (hotdog-2){\includegraphics[width=0.4cm, height=0.4cm]{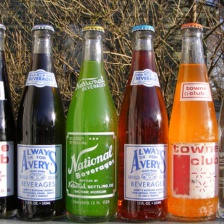}};
        \node (i_node1_1_2) [circle, minimum size=0.4cm, draw=light-orange, thick, below = 0.3cm of i_node1_1, xshift=0.5cm, label={[label distance=0cm, font=\bfseries, xshift=0cm]90: \scriptsize Eat}]{};
        \node (i_node1_1_2_1) [circle, minimum size=0.4cm, draw=light-orange, thick, below = 0.3cm of i_node1_1_2, xshift=-0.5cm, label={[label distance=0cm, font=\bfseries, xshift=0cm]-90: \scriptsize Corn}]{};
        \node (i_node1_1_2_2) [circle, minimum size=0.4cm, draw=light-orange, thick, below = 0.3cm of i_node1_1_2, xshift=0.3cm, label={[label distance=-0.2cm, font=\bfseries, xshift=0cm]45: \scriptsize Hotdog}]{};
        \node (hotdog-image-box) [rectangle, draw=black, below = 0.5cm of i_node1_1_2_2, minimum width = 1.5cm, minimum height=0.6cm]{};
        \node[inner sep=0pt, left of=hotdog-image-box, myshadow2, xshift=1.55cm, yshift=0cm] (hotdog-1){\includegraphics[width=0.4cm, height=0.4cm]{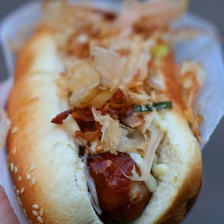}};
        \node[inner sep=0pt, right = 0.1cm of hotdog-1, myshadow2, label={[label distance=0.0cm]0: \small ...}] (hotdog-2){\includegraphics[width=0.4cm, height=0.4cm]{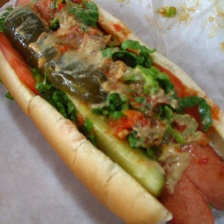}};
        \node (i_node1_2) [circle, minimum size=0.4cm, draw=light-orange, thick, below = 0.3cm of i_node1, xshift=-0.0cm, label={[label distance=0cm, font=\bfseries, xshift=0cm]0: \scriptsize Place}]{};
        \node (i_node1_2_1) [circle, minimum size=0.4cm, draw=light-orange, thick, below = 0.3cm of i_node1_2, xshift=0.5cm, label={[label distance=0cm, font=\bfseries, xshift=0cm]0: \scriptsize Nature}]{};
        \node (i_node1_2_1_1) [circle, minimum size=0.4cm, draw=light-orange, thick, below = 0.3cm of i_node1_2_1, xshift=0.3cm, label={[label distance=0cm, font=\bfseries, xshift=0cm]-90: \scriptsize Cliff}]{};
        \node (i_node1_2_1_2) [circle, minimum size=0.4cm, draw=light-orange, thick, below = 0.3cm of i_node1_2_1, xshift=-0.3cm, label={[label distance=0cm, font=\bfseries, xshift=0cm, text width=0.5cm]-90: \scriptsize Coral Reef}]{};
        \node (i_node1_3) [circle, minimum size=0.4cm, draw=light-orange, thick, below = 0.3cm of i_node1, xshift=2cm, label={[label distance=0cm, font=\bfseries, xshift=0cm]90: \scriptsize Species}]{};
        \node (i_node1_3_1) [circle, minimum size=0.4cm, draw=light-orange, thick, below = 0.3cm of i_node1_3, xshift=0.5cm, label={[label distance=0cm, font=\bfseries, xshift=0cm]60: \scriptsize Eukar.}]{};
        \node (i_node1_3_1_1) [circle, minimum size=0.4cm, draw=light-orange, thick, below = 0.3cm of i_node1_3_1, xshift=-0.25cm, label={[label distance=0cm, font=\bfseries, xshift=0.3cm]120: \scriptsize Animal}]{};
        \node (i_node1_3_1_1_1) [circle, minimum size=0.4cm, draw=light-orange, thick, below = 0.3cm of i_node1_3_1_1, xshift=-0.25cm, label={[label distance=0cm, font=\bfseries]100: \scriptsize Bird}]{};
        \node (bird-image-box) [rectangle, draw=black, below = 0.5cm of i_node1_3_1_1_1, minimum width = 1.5cm, minimum height=0.6cm]{};
        \node[inner sep=0pt, left of=bird-image-box, myshadow2, xshift=1.55cm, yshift=0cm] (bird-1){\includegraphics[width=0.4cm, height=0.4cm]{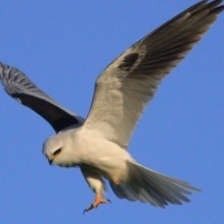}};
        \node[inner sep=0pt, right = 0.1cm of bird-1, myshadow2, label={[label distance=0.0cm]0: \small ...}] (bird-2){\includegraphics[width=0.4cm, height=0.4cm]{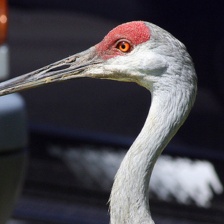}};
        \node (i_node1_3_1_1_2) [circle, minimum size=0.4cm, draw=light-orange, thick, below = 0.3cm of i_node1_3_1_1, xshift=0.25cm, label={[label distance=0cm, font=\bfseries, xshift=0.3cm]-90: \scriptsize Mamma}]{};
        \node (i_node1_3_1_2) [circle, minimum size=0.4cm, draw=light-orange, thick, below = 0.3cm of i_node1_3_1, xshift=0.5cm, label={[label distance=0cm, font=\bfseries, xshift=0cm]-90: \scriptsize Fungus}]{};
        \draw [arrow, dashed, draw=light-purple] (i_node1_1) -- (i_node1);
        \draw [arrow, dashed, draw=light-purple] (i_node1_1_1) -- (i_node1_1);
        \draw [arrow, dashed, draw=light-purple] (i_node1_1_1_1) -- (i_node1_1_1);
        \draw [arrow, dashed, draw=light-purple] (i_node1_1_2) -- (i_node1_1);
        \draw [arrow, dashed, draw=light-purple] (i_node1_1_2_1) -- (i_node1_1_2);
        \draw [arrow, dashed, draw=light-purple] (i_node1_1_2_2) -- (i_node1_1_2);
        \draw [arrow, dashed, draw=light-purple] (i_node1_2) -- (i_node1);
        \draw [arrow, dashed, draw=light-purple] (i_node1_2_1) -- (i_node1_2);
        \draw [arrow, dashed, draw=light-purple] (i_node1_2_1_1) -- (i_node1_2_1);
        \draw [arrow, dashed, draw=light-purple] (i_node1_2_1_2) -- (i_node1_2_1);
        \draw [arrow, dashed, draw=light-purple] (i_node1_3) -- (i_node1);
        \draw [arrow, dashed, draw=light-purple] (i_node1_3_1) -- (i_node1_3);
        \draw [arrow, dashed, draw=light-purple] (i_node1_3_1_1) -- (i_node1_3_1);
        \draw [arrow, dashed, draw=light-purple] (i_node1_3_1_2) -- (i_node1_3_1);
        \draw [arrow, dashed, draw=light-purple] (i_node1_3_1_1_1) -- (i_node1_3_1_1);
        \draw [arrow, dashed, draw=light-purple] (i_node1_3_1_1_2) -- (i_node1_3_1_1);
        \draw [dashed, draw=light-purple, thick] (i_node1_1_1_1) -- (beer-image-box);
        \draw [dashed, draw=light-purple, thick] (i_node1_1_2_2) -- (hotdog-image-box);
        \draw [dashed, draw=light-purple, thick] (i_node1_3_1_1_1) -- (bird-image-box);
    \end{tikzpicture}
    \end{adjustbox}
    \caption{(a) Examples of distribution shift in road sign, ImageNet, and Car Domains. 
    (b) Decomposition of elements w.r.t. the road signs and cars. It represents the association relations between object categories and object category elements.
    (c) Examples of the abstract representation of hierarchical relations in the ImageNet domain.
    }
    \label{fig:examples}
    \vspace{-1em}
\end{figure*}

%% file: sections/00_abstract.tex
\begin{abstract}
Despite the remarkable success of deep neural networks (DNNs) in computer vision, they fail to remain high-performing when facing distribution shifts between training and testing data.
In this paper, we propose \textbf{K}nowledge-\textbf{G}uided \textbf{V}isual representation learning (KGV) -- a distribution-based learning approach leveraging multi-modal prior knowledge -- to improve generalization under distribution shift.
It integrates knowledge from
two distinct modalities: 1) a knowledge graph (KG) with hierarchical and association relationships; and 2) generated synthetic images of visual elements semantically represented in the KG. 
The respective embeddings are generated from the given modalities in a common latent space, i.e., visual embeddings from original and synthetic images as well as knowledge graph embeddings (KGEs).
These embeddings are aligned via a novel variant of translation-based KGE methods, where the node and relation embeddings of the KG are modeled as Gaussian distributions and translations, respectively.
We claim that incorporating multi-model prior knowledge enables more regularized learning of image representations. 
Thus, the models are able to better generalize across different data distributions.
We evaluate KGV on different image classification tasks with major or minor distribution shifts, namely road sign classification across datasets from Germany, China, and Russia, image classification with the mini-ImageNet dataset and its variants, as well as the DVM-CAR dataset.
The results demonstrate that KGV consistently exhibits higher accuracy and data efficiency across all experiments. 
\end{abstract}

%% file: sections/01_introduction.tex
\section{Introduction}
Deep neural networks (DNNs) have demonstrated outstanding performance in computer vision (CV) tasks. 
They achieved remarkable success by leveraging vast amounts of data and computational power. 
However, these models often struggle with generalization, especially when the test data distribution deviates from the training data distribution~\cite{Liang2023ACS}. 
This happens as machine learning theory assumes that training and test data are identically distributed. Therefore, DNNs tend to overfit to the training domain.
The situation worsens under low data regimes \cite{adadi2021survey}, as DNNs are more prone to overfitting when insufficient training data exists.
Figure~\ref{fig:examples}a) illustrates three distinct examples of distribution shifts between train and test data, in particular the Road Sign, ImageNet and Car domain.
A DNN-based classifier applied to the testing domain will deteriorate in accuracy.
This happens because the visual elements of the images vary between the train and test domains. For example, in Germany, warning signs have a white background, whereas in China, they use a yellow background. Similarly, ImageNet and Car differentiate between train and test data because of their style.


Over the years, numerous approaches have been presented to utilize prior knowledge to enhance the performance of DNNs.
For instance, in the field of zero-shot learning (ZSL)~\citep{chen2023zero}, \citet{wang2018zero, Kampffmeyer_2019_CVPR} leverage additional information from a Knowledge Graphs (KG) to classify images of unseen classes. 
\citet{li2023knowledge} combines a knowledge-based classifier and a vision-based classifier to enhance the model's ability to classify new categories.  
These methods are excepted to classify unseen classes based on the other seen classes sharing the same superclass. 
A concrete example is the classification of unseen zebras based on seen \emph{horses} and \emph{donkeys} as they both belong to the same superclass, i.e., \emph{equus}. 
The prior knowledge contained in their KG pertains to the hierarchical levels of biological classification. 

However, approaches that only use symbolic knowledge from the KG are limited to addressing the issue of data distribution shift as the symbolic knowledge lacks the connection to the visual domain. 
We hypothesize that the model's ability to generalize can be further improved by leveraging multi-modal prior knowledge that (partially) covers new test data distributions. The multi-model prior knowledge here indicates both a KG with hierarchical and association relations and generated synthetic images of visual elements semantically represented in the KG. 
To investigate this hypothesis, in this paper, we propose a novel neuro-symbolic approach, which aligns the image embeddings and knowledge graph embeddings (KGEs) in a common latent space by means of a variant of translation-based KGE method~\cite{NIPS2013_1cecc7a7}.
Thus, it regularizes the latent space with the guidance of the multi-modal prior knowledge to prevent the model from overfitting. 


First, we capture information about the structures of objects in a KG~\cite{hogan2021knowledge} to inform and enhance the learning process. 
For instance, warning signs are composed of a white triangle with a red border plus one pictographic element (cf.\ Figure~\ref{fig:examples}b).
Road signs in different countries share the same elements, such as the shape or the pictographic element, or both.
Also, the hierachical relations are demonstrated in Figure~\ref{fig:examples}c. 
Combining this information, we construct a KG and develop a translation-based KGE method to map the nodes of the KG to embeddings in the latent space. 
An object category (e.g., `danger sign') is naturally represented by its set of image vectors in the latent space. 
To align image and object categories, we represent the latter by a Gaussian distribution in the common latent space. Thus, the embeddings of object category elements and their relations to object categories indirectly influence the formation of categories in the latent space. 


Second, for each object category element, we generate their respective synthetic images (if applicable) to augment the dataset. 
For instance, in the road sign domain, these elements are shape, color, and legend within the sign. 
These images are considered as visual prior knowledge of the object categories defined in the KG. 
Therefore, in this paper, the prior knowledge we used has two distinct modalities: 1) the factual knowledge across different data distributions represented in a KG, and 2) generated synthetic images of object category elements aligned with their semantic representation in the KG. 

We evaluate the performance of our approach under distribution shifts and low data regimes in the context of image classification. 
To represent the prior knowledge, we construct three targeted KGs for the domains of Road Sign Recognition, Car Recognition, and ImageNet. 
Our experimental results show that KGV outperforms the image classification baselines in all experiments. 
Additionally, we found that our method is more data-efficient.
Our main contributions are summarized as follows:
\begin{itemize}
    \item We propose our KGV method, an end-to-end neuro-symbolic approach that uses multi-modal prior knowledge to regularize the latent space and thus enhance the generalization ability under data distribution shift. 
    \item We developed a new variant of translation-based KGE to align the embeddings of images and the KG in a common embedding space, where the nodes and predicates in the KG are represented as \emph{Gaussian distributions} and \emph{translations}, respectively. 
    \item We evaluate KGV  on various image classification datasets, namely road sign datasets, mini-ImageNet and its variants, as well as the DVM-CAR datasets. The results illustrate a notable improvement under distribution shift: 4.4\% and 4.1\% on average in the ImageNet and Road Sign Recognition domains.  Moreover, our method improves the SOTA results on the DVM-CAR dataset and shows significant improvements on data efficiency in low data regimes.  
    \item We analyse whether KGV can be adapted to the SOTA vision foundation models like CLIP and DINOv2. 
    The results indicate that integrating KGV, CLIP and DINOv2 exhibit a better performance across various datasets.  
    The code base is available at \url{https://anonymous.4open.science/r/kgv-D68F/}.
\end{itemize}

%% file: sections/02_related_works.tex
\section{Preliminaries}
\label{sec:preliminaries}
\subsection{Visual Embeddings}

A visual encoder network $f_{\bs{\phi}}(\cdot): \mathcal{I}\rightarrow \mathbb{R}^d$ with parameters of $\bs{\phi}$ maps images in image space $\mathcal{I}$ to a visual embeddings $\bs{z^I} \in \mathbb{R}^d$, where $d$ is the dimension of the latent space. 
This function is typically realized through a series of transformations involving neural networks, particularly convolution-based models \cite{DBLP:journals/corr/HeZRS15}, or Transformer-based models~\cite{dosovitskiy2021image, carion2020endtoend}. 

\subsection{Knowledge Graph Embeddings}

KG stores factual information of real-world entities and their relationships in triple format $\langle \textit{head}, \textit{predicate}, \textit{tail}\rangle$.
Knowledge graph embedding (KGE) aims to represent entities and predicates in KG into low-dimensional vectors while preserving their semantic and structural information. 


Formally, we define a KG as $\mathcal{G}\subseteq \mathcal{E}\times \mathcal{R}\times \mathcal{E}$ over a set $\mathcal{E}$ of entities and a set $\mathcal{R}$ of predicates. 
A KGE model maps entities and predicates into $d$-dimensional embedding $M_\theta: \mathcal{E}\cup \mathcal{R}\rightarrow \mathbb{R}^d$.  
We denote $\langle h,r,t\rangle$ as a triple in $\mathcal{G}$. 
A score function $s(\textbf{h},\textbf{r},\textbf{t})$ is defined on the embeddings of a triple, which should assign higher scores to positive triples, i.e., true facts in the KG, and lower scores to negative triples, i.e., corrupted triples, generated by 
randomly replacing the head or tail entity in an observed triple with a random entity sampled from $\mathcal{E}$. The parameter $\theta$ is learned by minimizing cross-entropy loss~\cite{NIPS2013_1cecc7a7} or margin-based loss~\cite{sun2019rotate}.



\subsection{Few-shot Learning}
A classifier $f: x \rightarrow y$ is initially trained with a set of labeled training samples $\mathcal{D}_{\textrm{tr}}=\{(x,y)|x \in \mathcal{X}, y \in \mathcal{Y}\}$. 
Few-shot learning focuses on a common scenario called $N$-way, $K$-shot learning. Given a set of $K$ labeled examples for each of $N$ different classes $\mathcal{D}_{\textrm{few}}=\{(x,y)|x \in \mathcal{X}_{\textrm{few}}', y \in \mathcal{Y'}\}$, where $|\mathcal{Y'}|=N$ and $\forall y_0 \in \mathcal{Y'}, |\{(x,y) \in \mathcal{D}_\textrm{few}| y = y_0\}| = K$, in the target dataset, this classifier is able to correctly predict the labels of samples in the full target dataset $\mathcal{D}_{\textrm{full}}=\{(x,y)|x \in \mathcal{X}_{\textrm{full}}', y \in \mathcal{Y'}\}$, with $(\mathcal{X} \cup \mathcal{X}_{\textrm{few}}') \cap \mathcal{X}_{\textrm{full}}' = \emptyset$.

\section{Related Work}
\label{sec:related works}
\subsection{Use Prior Knowledge for Learning Tasks}
Implicit prior knowledge in pre-trained models like CLIP~\cite{radford2021learning} and DINOv2~\cite{dinov2} can be leveraged by continuing to fine-tune them.
Including explicit knowledge from KGs can also enhance model's ability in visual or textual tasks~\cite{DBLP:journals/semweb/MonkaHR22}. 
Some approaches~\cite{annervaz2018learning, bosselut-etal-2019-comet, liu2020k} leverage KGs to enhance language models with better reasoning ability. In the field of computer vision, KGs also play an important role. \citet{NEURIPS2023_2b25c397} introduce a KG that includes inter-class relationships to the visual-language model, yielding a more effective classifier for downstream tasks. Some approaches ~\cite{zareian2020bridging, yu2021ernie} build scene graphs to enhance their models' performance. 
There also exist some approaches use extra information in the KG to perform zero-shot learning tasks \cite{wang2018zero, Kampffmeyer_2019_CVPR, DBLP:journals/corr/abs-2012-06236} or few-shot learning tasks \cite{10038499, DBLP:journals/corr/abs-2001-08735}. 

In the field of symbolic AI, some approaches include textual information \cite{nayyeri2023integrating}, or multi-modal information \cite{chen2022hybrid, liang2023structure}, to increase the knowledge graph reasoning ability. They combine the structured information from KGs, visual information, and textual information with transformer-based approaches. These works prove that including visual or textual information or both is helpful for knowledge graph based reasoning and knowledge graph completion. 

\subsection{Integrate Knowledge Graphs with Other Modalities}
Two primary approaches are recognized in the literature to encapsulate the methodologies for integrating heterogeneous information from KGs with other modalities, such as visual and textual information. The first category \cite{chen2022hybrid, liang2023structure, mousselly2018multimodal} involves merging structural data derived from KGs — typically obtained via Graph Neural Networks (GNNs)—through the utilization of cross-attention mechanisms as employed in Transformer models. The second category \cite{NEURIPS2023_2b25c397, 10.1007/978-3-030-88361-4_21, DBLP:conf/semweb/MonkaHR22} entails aligning KGEs with visual or textual embeddings in one common space by contrastive learning \cite{radford2021learning}. 


In this paper, we propose a novel approach that leverages the multi-modal prior knowledge from both, a KG and synthetic images generated based on visual features that are semantically represented in the KG, in order to boost the model's ability to generalize under data distribution shifts. Our approach aligns KGEs and visual embeddings in the \emph{same} latent space using a new variant of translation-based KGE methods, where the nodes and relations are represented as \emph{Gaussian distributions} and \emph{translations}, respectively, to model the structured information in the KG within the latent space. Existing translation-based KGE methods typically represent entities and predicates with vector embeddings \cite{NIPS2013_1cecc7a7, wang2014knowledge, lin2015learning, socher2013reasoning}. The authors \cite{he2015learning, xiao2015transg} take the uncertainty of entities and predicates into account and represent them by probability distributions.

%% file: sections/03_methodology.tex
\section{Methodology}
\setlength{\headheight}{15.45749pt}
\label{sec:methodology}



\subsection{Overview}

We propose KGV, an approach to address the limitations of the deep learning solutions for supervised image classification and few-shot learning w.r.t. to the distribution shifts.
The core idea of KGV is the inclusion of multi-modal prior knowledge coming from two different modalities: 1) domain knowledge - captured in a semantic KG; and 2) synthetic images - generated for object category elements.
We encode the node and relation information from the KG and images as KGEs ($ \boldsymbol{r}, \boldsymbol{t}$) and image embeddings $\boldsymbol{z^I}$ into one latent space and align them via a variant of translation-based KGE method.
Here, we represent the object categories in KG as Gaussian embeddings in the latent space, and we assume each image embedding should follow the distribution of its corresponding category distribution.
It extends the contrastive learning, as seen in SupCon \cite{DBLP:journals/corr/abs-2004-11362} and  CLIP~\cite{radford2021learning}, which focuses solely on similarity and dissimilarity, by incorporating the capability to model additional individual relationships between entities.

Our approach is designed end-to-end by adding the regularization loss and cross-entropy loss together to train the network. The regularization loss aligns the KG embedding and the visual embedding. The cross-entropy loss is used for the image classification. 
The architecture of KGV is depicted in Figure~\ref{fig:architeture}. 
Our model consists of three phases: the knowledge modeling phase, the training phase, and the inference phase. 
Here, we take the \emph{road sign domain} as an example to describe our model.

\subsection{Knowledge Modeling}
Prior knowledge used in our KGV exists in two distinct modalities.
Firstly, an expert-constructed KG includes factual knowledge such as hierarchical and association relations across different data distributions. 
Secondly, we generate synthetic images of object category elements that are semantically represented in the KG but lack visual information in the image dataset. In the road sign recognition domain, these elements are shape, color, and the sign legend. 

\subsubsection{Knowledge Graph Construction:}

\input{tikz/architecture2}
We construct a KG to represent object categories, object category elements, and the relationships between an object category and its elements based on domain knowledge. 
At the top level, there are two different categories: \emph{road sign} and \emph{road sign feature}. 
The \emph{road sign} category comprises more sub-categories: \emph{informative}, \emph{prohibitory}, \emph{mandatory}, and \emph{warning}. 
Further, we define dedicated classes, such as Germany, China, and Russia, to model various country versions of the road signs. 
The \emph{road sign feature} category contains sub-categories of \emph{shape}, \emph{color}, and \emph{legend within the sign}. 
An illustration of hierarchical relations is demonstrated in the appendix. 
Besides the hierarchical information, the KG also contains four association relations. 
They are `has the shape of', `has the sign legend of', `has the background color of', and `has the border color of'.
For instance, as Figure~\ref{fig:examples}b shows, the danger sign has the shape of a triangle and has the sign legend of the exclamation mark. 

\subsubsection{Synthetic images of object category elements:}
We generate synthetic images of object category elements such as shapes and colors, which is treated as visual prior knowledge. 
In our setting, images of different colors are generated by randomly selecting an image within the color range corresponding to the popular color definition. 
The images of shapes are generated by alternating on the thickness, size, and location on the canvas. 
The images of sign legends are extracted from standard road signs, then randomly resized and positioned at various locations on the canvas. 
These synthetic images are then added to the road sign dataset and trained together with those road sign images. All types of synthetic images can be seen in the appendix. 

\subsection{Training}
Our training loss comprises two components: the regularization loss, which is used to align the image embeddings with the KGEs, and the cross-entropy loss, which is employed for classification purposes. 
The regularization loss is specifically formulated to maximize the score function value for positive triplets and minimize it for negative triplets. In the following subsections, we first delineate the image, node, and relation embeddings. 
Subsequently, we elucidate the methodology for constructing triplets of embeddings and the process for masking the positive and negative triplets. 
Next, we introduce the design of the score function for triplets. Finally, we provide a summary of the overall loss function.

\subsubsection{Embeddings:}
Each image is provided as input into the image encoder to obtain the respective image embedding $\bs{z^{I}} \in \mathbb{R}^d$ in the latent space, where $d$ is the size of the dimension.
The relations from KG connecting object categories, their elements and their respective images are encoded and stored in the relation lookup table, which is represented as a list of embeddings $[\bs{z_0^{r}}, \bs{z_1^{r}}, ..., \bs{z_{N_r}^r}]$, where each $\bs{z_i^{r}}$ is the $i$-th relation vector embedding with the size of $d$. 
Note that the 0-th relation represents the inclusion relation demonstrated with purple dashed arrows as illustrated in Figure \ref{fig:architeture}. 
The hierarchy of road signs is preserved in the KG.
For instance, the images of the danger signs in Germany belong to the German danger sign category, and they also belong to the danger sign category, warning sign category, and road sign category. 
We use $\bs{z_0^r}$ to represent all these inclusion relations. 
In terms of the other $N_r$ relations, if one node $h$ has one relation with another node $t$, then all the sub-category nodes of $h$ have such relation with the node $t$.
For example, all warning signs have a triangle shape, including their sub-categories like danger and danger animal.
The node lookup table is represented as a list of embeddings $[\bs{z_1^{o}}, \bs{z_2^{o}}, ..., \bs{z_{N_o}^o}]$, where each $\bs{z_j^{o}}$ denotes the $j$-th  embedding. $N_o$ is the total number of nodes in the KG. 
The node embeddings are considered as Gaussian embeddings $\bs{z_j^o} = (\bs{\mu_j}, \bs{\Sigma_j})$, where $\bs{\mu_j}$ and $\bs{\Sigma_j}$ are the mean and variance of the Gaussian distribution.
Note that nodes, their relations, and respective images are embedded in the same latent space.

\subsubsection{Positive and Negative Triplets of Embeddings:}
Following the close world assumption in the context of KG, any triplet not explicitly defined in the KG is considered to be a negative one. 
Since $N_o$ nodes and $N_r+1$ relations exist in the KG, for each image embedding $\bs{z^I}$, $(N_r+1) \times N_o$ triplets with the embedding form of $\langle \bs{z^I}, \bs{z^r_i}, \bs{z^o_j}\rangle$ can be created. 
The positive triplets represent the defined information in the KG about the given image. 
To mask all the positive triplets and negative triplets for one given image, we construct a tensor $\bs{M} \in [0,1]^{(N_r+1) \times N_o}$:
\begin{equation}
    \label{eq:mask}
    \bs{M}_{ij} = 
    \begin{cases}
        1 & \text{if } \langle e_{node(y)}, r_i, e_j \rangle \in \mathcal{G} \\
        1 & \text{if } r_i = \text{instanceOf} \wedge j=node(y) \\
        0 & \text{otherwise}
    \end{cases}
\end{equation}
where $e_j$ represents the $j$-th element in entity set $\mathcal{E}$. $r_i$ is the $i$-th element in the relation set $\mathcal{R}$. $\mathcal{G}$ represents the triplet stored in the KG. 
$y$ is the class label of the given image. $node(\cdot)$ indicates the mapping function connecting the class id of the image to the corresponding node id in the KG. 
As Figure \ref{fig:architeture} shows, positive triplets are marked with green color, whereas negative triplets are marked with white color. 

\subsubsection{Score Function Design:}
For a given triplet $\langle \bs{z^I}, \bs{z_i^r}, \bs{z_j^o} \rangle$, we design the score function in translation-based manner which means $\bs{z^I} + \bs{z_i^r} \approx \bs{z_j^o}$.
In this paper, we represent image embeddings and relation embeddings as vector embeddings. The node embeddings are considered as Gaussian embeddings $\bs{z_j^o} = (\bs{\mu_j}, \bs{\Sigma_j})$ to better handle the inclusion relationship between the image embedding and the node embedding, where $\bs{\mu_j}$ and $\bs{\Sigma_j}$ are the mean and variance of the Gaussian distribution. The intuition behind this is that each node in the KG contains various image instances. Thus, it is more reasonable to model the node embedding as Gaussian since each node represents a cluster of image vector embeddings. The score function can be defined as:
\begin{equation}
    \label{eq:score function}
    s(\bs{z^I}, \bs{z_i^r}, \bs{z_j^o} ) = \mathcal{N}(\bs{z^I} + \bs{z_i^r}; \bs{\mu_j}, \bs{\Sigma_j})
\end{equation}
where $\mathcal{N}(\bs{z^I} + \bs{z_i^r}; \bs{\mu_j}, \bs{\Sigma_j})$ denotes the possibility density of the vector $\bs{z^I} + \bs{z_i^r}$ under the Gaussian distribution with parameters of $\bs{\mu_j}$ and $\bs{\Sigma_j}$. Different score function designs are also compared in our experiments.

\subsubsection{Loss Functions:}
During the training phase, there are two losses to minimize. The first one is the cross-entropy loss $\mathcal{L}_\textrm{CE}(\bs{\hat{y}}, \bs{y})$ to classify the images to their corresponding classes, where $\bs{\hat{y}}$ is the predicted possibility of classes and $\bs{y}$ is the ground truth class label in the form of one-hot-key.

The other loss term is the regularization loss. It aims to regularize the latent space with prior knowledge by aligning image embeddings with KG embeddings. 
Here, we represent the relationships from the KG in the visual latent space by optimizing the score function as defined in Equation \eqref{eq:score function}. 
If the score function has high scores for positive triplets and low scores for negative triplets, the image embeddings and the KGEs are then well aligned in the latent space. 
The regularization loss is defined as follows:


\begin{equation}
    \label{eq:regularization loss}
    \mathcal{L}_{\textrm{reg}} = \frac{\sum_{i,j} \bs{M}_{ij} \cdot  \bs{S}_{ij}}{\sum_{i,j} \bs{M}_{ij}} + \frac{\sum_{i,j} (1-\bs{M}_{ij}) \cdot max\{0, \epsilon - \bs{S}_{ij}\}}{\sum_{i,j} 1 - \bs{M}_{ij}}
\end{equation}

where $\textbf{M}$ is the mask defined in Equation \eqref{eq:mask}. $\textbf{S} \in R^{(N_r+1) \times N_o}$ and $\textbf{S}_{ij} = - \log(s(\bs{z^I}, \bs{z_i^r}, \bs{z_j^o}))$, following the negative log likelihood form. $\epsilon$ is the threshold score difference between positive and negative triplets.
Finally, the total loss is defined by:
\begin{equation}
    \label{eq:total loss}
    \mathcal{L} = \mathcal{L}_{\textrm{CE}} + \beta \mathcal{L}_{\textrm{reg}}\text{,}
\end{equation}
where $\beta$ is the hyper-parameter used to balance the cross-entropy loss term and regularization loss term. 
\subsection{Inference Phase}
At the inference phase, the classification task is completed by selecting the class with the highest probability based on the output of the decoder $\bs{\hat{y}}$. 

%% file: tikz/architecture2.tex
\tikzstyle{arrow} = [thick,->,>=stealth]
\tikzstyle{myshadow} = [drop shadow={
            shadow scale=0.95,
            shadow xshift=0.5ex,
            shadow yshift=-0.5ex
        }]
\tikzstyle{myshadow2} = [drop shadow={
            shadow scale=0.95,
            shadow xshift=0.75ex,
            shadow yshift=-0.75ex
        }]
\tikzstyle{encoder} = [rectangle, rounded corners, text centered, draw=black, fill=network-blue!70, minimum height=2cm, minimum width=3cm]

\begin{figure*}[ht]
    \centering
    \begin{adjustbox}{width=1.0\textwidth}
    \begin{tikzpicture}[node distance=2cm]
        \node (phase1) [rectangle, rounded corners, fill=light-red!20, minimum height=7.8cm, minimum width=24cm, label={[label distance=-0.75cm,font=\bfseries, xshift=-8.9cm]90: \Large 1. Knowledge Modelling Phase}] {};

        \node(images) [rectangle, minimum height=6.5cm, minimum width=7.5cm,left of=phase1, xshift=-5.75cm, yshift=-0.25cm, draw=black, thick, label={[label distance=-0.5cm, font=\bfseries, xshift=2.2cm]-90: Road Sign Domain}]{};
        \node (color-images) [rectangle, minimum height=2cm, minimum width=2.25cm,  left of=images, xshift=-0.5cm, yshift=1.5cm, label={[label distance=0cm, font=\bfseries]90: Colors}]{};
        \node[inner sep=0pt, above left of=color-images, xshift=0.9cm, yshift=-0.8cm, line width=3pt, myshadow] (color-yellow){\includegraphics[width=0.75cm, height=0.75cm]{images/Color_Yellow.png}};
        \node[inner sep=0pt, right = 0.2cm of color-yellow, myshadow] (color-red){\includegraphics[width=0.75cm, height=0.75cm]{images/Color_Red.png}};
        \node[inner sep=0pt, below = 0.2cm of color-yellow, myshadow, label={[label distance=0.25cm]0: \Large ...}] (color-blue){\includegraphics[width=0.75cm, height=0.75cm]{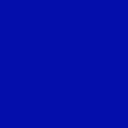}};

        \node (shape-images) [rectangle, minimum height=2cm, minimum width=2.25cm,  right = 0.2cm of color-images, label={[label distance=0cm, font=\bfseries]90: Shapes}]{};
        \node[inner sep=0pt, above left of=shape-images, xshift=0.9cm, yshift=-0.8cm, myshadow] (shape-circle){\includegraphics[width=0.75cm, height=0.75cm]{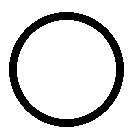}};
        \node[inner sep=0pt, right = 0.2cm of shape-circle, myshadow] (shape-diamond){\includegraphics[width=0.75cm, height=0.75cm]{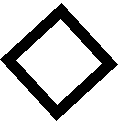}};
        \node[inner sep=0pt, below = 0.2cm of shape-circle, myshadow,label={[label distance=0.25cm]0: \Large ...}] (shape-triangleUp){\includegraphics[width=0.75cm, height=0.75cm]{images/Shape_TriangleUp.png}};

        \node (icon-images) [rectangle, minimum height=2cm, minimum width=2.25cm,  right = 0.2cm of shape-images, label={[label distance=0cm, font=\bfseries]90: Sign Legends}]{};
        \node[inner sep=0pt, above left of=icon-images, xshift=0.9cm, yshift=-0.8cm, myshadow] (icon-bicycle){\includegraphics[width=0.75cm, height=0.75cm]{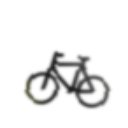}};
        \node[inner sep=0pt, right = 0.2cm of icon-bicycle, myshadow] (icon-animal){\includegraphics[width=0.75cm, height=0.75cm]{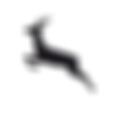}};
        \node[inner sep=0pt, below = 0.2cm of icon-bicycle, myshadow, label={[label distance=0.25cm]0: \Large ...}] (icon-exclamation){\includegraphics[width=0.75cm, height=0.75cm]{images/Icon_Exclamation.png}};

        \node (roadsign-images) [rectangle, minimum height=4.5cm, minimum width=6.5cm, below = 1.25cm of shape-images, yshift=0.5cm, label={[label distance=-0.5cm, font=\bfseries, xshift=-0.6cm]120: Road Signs}]{};
        \node[inner sep=0pt, above left of=roadsign-images, xshift=-1.5cm, yshift=0cm, myshadow] (rs-goleftorstraight){\includegraphics[width=0.75cm, height=0.75cm]{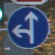}};
        \node[inner sep=0pt, right = 0.2cm of rs-goleftorstraight, myshadow] (rs-speedlimit30){\includegraphics[width=0.75cm, height=0.75cm]{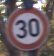}};
        \node[inner sep=0pt, right = 0.2cm of rs-speedlimit30, myshadow] (rs-giveway){\includegraphics[width=0.75cm, height=0.75cm]{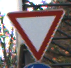}};
        \node[inner sep=0pt, right = 0.2cm of rs-giveway, myshadow] (rs-stop){\includegraphics[width=0.75cm, height=0.75cm]{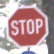}};
        \node[inner sep=0pt, right = 0.2cm of rs-stop, myshadow] (rs-speedlimit120){\includegraphics[width=0.75cm, height=0.75cm]{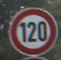}};
        \node[inner sep=0pt, right = 0.2cm of rs-speedlimit120, myshadow] (rs-roundabout){\includegraphics[width=0.75cm, height=0.75cm]{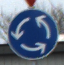}};
        \node[inner sep=0pt, right = 0.2cm of rs-roundabout, myshadow] (rs-roadconstruction){\includegraphics[width=0.75cm, height=0.75cm]{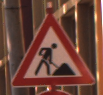}};
        \node[inner sep=0pt, below = 0.2cm of rs-goleftorstraight, myshadow] (rs-trafficlight){\includegraphics[width=0.75cm, height=0.75cm]{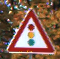}};
        \node[inner sep=0pt, right = 0.2cm of rs-trafficlight, myshadow] (rs-priority){\includegraphics[width=0.75cm, height=0.75cm]{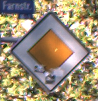}};
        \node[inner sep=0pt, right = 0.2cm of rs-priority, myshadow] (rs-dangerintersection){\includegraphics[width=0.75cm, height=0.75cm]{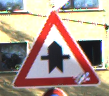}};
        \node[inner sep=0pt, right = 0.2cm of rs-dangerintersection, myshadow] (rs-danger){\includegraphics[width=0.75cm, height=0.75cm]{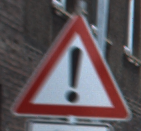}};
        \node[inner sep=0pt, right = 0.2cm of rs-danger, myshadow] (rs-animal){\includegraphics[width=0.75cm, height=0.75cm]{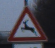}};
        \node[inner sep=0pt, right = 0.2cm of rs-animal, myshadow, label={[label distance=0.25cm]0: \Large ...}] (rs-endrestriction){\includegraphics[width=0.75cm, height=0.75cm]{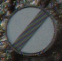}};

        \node (knowledge-graph) [rectangle, minimum height=7cm, minimum width=14.75cm, right = 0.75cm of images, yshift=0.25cm, draw=black, thick, label={[label distance=-0.8cm,  font=\bfseries]45: Visual Knowledge Graph}]{};
        \draw [-{Triangle Cap[]. Fast Triangle[] Fast Triangle[]}, draw=black, line width=1mm] (images.east) -- (images.east-|knowledge-graph.west);
        \node (road-sign-node) [circle, minimum size=0.6cm,  very thick, draw=light-orange,  left of = knowledge-graph, xshift=-2.25cm, yshift=2.5cm, label = {[label distance=-0.1cm, font=\bfseries]90: \footnotesize Road Sign}]{};
        \node (warning-node) [circle, minimum size=0.6cm,  very thick, draw=light-orange,  below=0.5cm of road-sign-node, xshift=0.75cm, label = {[label distance=-0.1cm, font=\bfseries]80: \footnotesize Warning}]{};
        \node (mandatory-node) [circle, minimum size=0.6cm,  very thick, draw=light-orange,  below=0.5cm of road-sign-node, xshift=-1.25cm, label = {[label distance=-0.1cm, font=\bfseries]93: \footnotesize Mandatory}]{};
        \node (ellipsis-road-sign)[right = 0.4cm of mandatory-node]{\Large ...};
        \node (ellipsis-mandatory)[below = 0.4cm of mandatory-node, xshift=-0.7cm]{\Large ...};
        \node (danger-animal-node) [circle, minimum size=0.6cm,  very thick, draw=light-orange,  below=0.5cm of warning-node, xshift=-1cm, label = {[label distance=-0.1cm, font=\bfseries, text width=0.7cm]100: \footnotesize Danger\\Animal}]{};
        \node (ellipsis-danger-animal)[below = 0.4cm of danger-animal-node, xshift=-0.7cm]{\Large ...};
        \node (ellipsis-warning)[right = 0.1cm of danger-animal-node]{\Large ...};
        \node (danger-sign-node) [circle, minimum size=0.6cm,  very thick, draw=light-orange,  below=0.5cm of warning-node, xshift=0.5cm, label = {[label distance=-0.1cm, font=\bfseries, text width=1.0cm]40: \footnotesize Danger\\Sign}]{};
        \node (danger-sign-de-node) [inner sep=0pt, circle, minimum size=0.6cm,  very thick, draw=light-orange,  below=0.5cm of danger-sign-node, xshift=-0.5cm]{\footnotesize \textbf{DE}};
        \node (danger-sign-cn-node) [inner sep=0pt, circle, minimum size=0.6cm,  very thick, draw=light-orange,  below=0.5cm of danger-sign-node, xshift=0.5cm]{\footnotesize \textbf{CN}};
        \node (icon-node) [circle, minimum size=0.6cm,  very thick, draw=light-orange,  right of = knowledge-graph, xshift=0.75cm, yshift=-0.5cm, label = {[label distance=0cm, font=\bfseries]90: \footnotesize Sign Legends}]{};
        \node (icon-exclamation-node) [circle, minimum size=0.6cm,  very thick, draw=light-orange,  below=0.5cm of icon-node, xshift=-0.75cm, label = {[label distance=0cm, font=\bfseries]-90: \footnotesize Exclamation}]{};
        \node (ellipsis-icon)[right = 0.2cm of icon-exclamation-node]{\Large ...};
        \node (ellipsis-exclamation-instances)[below = 0.2cm of icon-exclamation-node, xshift=-1cm]{\Large ...};
        \draw [dashed, draw=light-purple, very thick] (icon-exclamation-node) -- +(-1cm, 0) -- (ellipsis-exclamation-instances);
        \node (icon-animal-node) [circle, minimum size=0.6cm,  very thick, draw=light-orange,  below=0.5cm of icon-node, xshift=0.75cm, label = {[label distance=-0.1cm, font=\bfseries]70: \footnotesize Animal}]{};

        \node (shape-node) [circle, minimum size=0.6cm,  very thick, draw=light-orange,  right of = knowledge-graph, xshift=0cm, yshift=2.5cm, label = {[label distance=0cm, font=\bfseries]90: \footnotesize Shapes}]{};
        \node (shape-triangle-node) [circle, minimum size=0.6cm,  very thick, draw=light-orange,  below=0.5cm of shape-node, xshift=-0.75cm, label = {[label distance=-0.1cm, font=\bfseries]100: \footnotesize Triangle}]{};
        \node (ellipsis-shapes)[right = 0.2cm of shape-triangle-node]{\Large ...};
        \node (ellipsis-triangle-instances)[below = 0.6cm of shape-triangle-node]{\Large ...};
        \draw [dashed, draw=light-purple, very thick] (shape-triangle-node) -- (ellipsis-triangle-instances);
        \node (shape-circle-node) [circle, minimum size=0.6cm,  very thick, draw=light-orange,  below=0.5cm of shape-node, xshift=0.75cm, label = {[label distance=-0.1cm, font=\bfseries]80: \footnotesize Circle}]{};

        \draw [arrow, draw=light-purple] (danger-sign-node) to node [above,midway,, rotate=-22, color=light-purple] {hasSignLegendOf} (icon-exclamation-node);
        \draw [arrow, draw=light-purple] (warning-node) to node [above,midway, rotate=0, color=light-purple] {hasShapeOf} (shape-triangle-node);
        \draw [arrow, dashed, draw=light-purple] (danger-sign-de-node) -- (danger-sign-node);
        \draw [arrow, dashed, draw=light-purple] (danger-sign-cn-node) -- (danger-sign-node);
        \draw [arrow, dashed, draw=light-purple] (danger-sign-node) -- (warning-node);
        \draw [arrow, dashed, draw=light-purple] (ellipsis-danger-animal) -- (danger-animal-node);
        \draw [arrow, dashed, draw=light-purple] (warning-node) -- (road-sign-node);
        \draw [arrow, dashed, draw=light-purple] (ellipsis-mandatory) -- (mandatory-node);
        \draw [arrow, dashed, draw=light-purple] (mandatory-node) -- (road-sign-node);
        \draw [arrow, dashed, draw=light-purple] (danger-animal-node) -- (warning-node);
        \draw [arrow, dashed, draw=light-purple] (icon-exclamation-node) -- (icon-node);
        \draw [arrow, dashed, draw=light-purple] (icon-animal-node) -- (icon-node);
        \draw [arrow, dashed, draw=light-purple] (shape-triangle-node) -- (shape-node);
        \draw [arrow, dashed, draw=light-purple] (shape-circle-node) -- (shape-node);


        \node (danger-sign-image-box) [rectangle, draw=black, very thick, left of =knowledge-graph, minimum width = 2cm, minimum height=2cm, xshift=-3.75cm, yshift=-2.25cm]{};
        \node[inner sep=0pt, left of=danger-sign-image-box, myshadow, xshift=1.5cm, yshift=0.5cm] (rs-animal-1){\includegraphics[width=0.75cm, height=0.75cm]{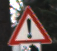}};
        \node[inner sep=0pt, right = 0.2cm of rs-animal-1, myshadow] (rs-animal-2){\includegraphics[width=0.75cm, height=0.75cm]{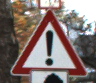}};
        \node[inner sep=0pt, below = 0.2cm of rs-animal-1, myshadow, label={[label distance=0.25cm]0: \Large ...}] (rs-animal-3){\includegraphics[width=0.75cm, height=0.75cm]{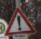}};

        \node (danger-sign-image-box_cn) [rectangle, draw=black, very thick, left of =knowledge-graph, minimum width = 2cm, minimum height=2cm, xshift=0.75cm, yshift=-2.25cm]{};
        \node[inner sep=0pt, left of=danger-sign-image-box_cn, myshadow, xshift=1.5cm, yshift=0.5cm] (rs-danger-cn-1){\includegraphics[width=0.75cm, height=0.75cm]{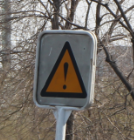}};
        \node[inner sep=0pt, right = 0.2cm of rs-danger-cn-1, myshadow] (rs-danger-cn-2){\includegraphics[width=0.75cm, height=0.75cm]{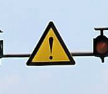}};
        \node[inner sep=0pt, below = 0.2cm of rs-danger-cn-1, myshadow, label={[label distance=0.25cm]0: \Large ...}] (rs-danger-cn-3){\includegraphics[width=0.75cm, height=0.75cm]{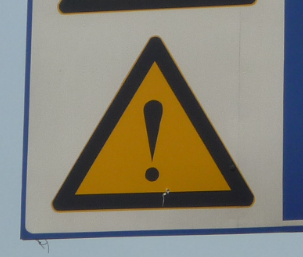}};
        
        \node (icon-animal-image-box) [rectangle, draw=black, very thick, right of =knowledge-graph, minimum width = 2cm, minimum height=2cm, xshift=3.75cm, yshift=-1cm]{};
        \node[inner sep=0pt, left of=icon-animal-image-box, myshadow, xshift=1.5cm, yshift=0.5cm] (icon-animal-1){\includegraphics[width=0.75cm, height=0.75cm]{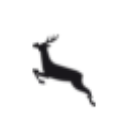}};
        \node[inner sep=0pt, right = 0.2cm of icon-animal-1, myshadow] (icon-animal-2){\includegraphics[width=0.75cm, height=0.75cm]{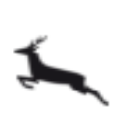}};
        \node[inner sep=0pt, below = 0.2cm of icon-animal-1, myshadow, label={[label distance=0.25cm]0: \Large ...}] (icon-animal-3){\includegraphics[width=0.75cm, height=0.75cm]{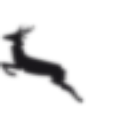}};

        \node (shape-circle-image-box) [rectangle, draw=black, very thick, right of =knowledge-graph, minimum width = 2cm, minimum height=2cm, xshift=2.75cm, yshift=1.75cm]{};
        \node[inner sep=0pt, left of=shape-circle-image-box, myshadow, xshift=1.5cm, yshift=0.5cm] (shape-circle-1){\includegraphics[width=0.75cm, height=0.75cm]{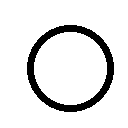}};
        \node[inner sep=0pt, right = 0.2cm of shape-circle-1, myshadow] (shape-circle-2){\includegraphics[width=0.75cm, height=0.75cm]{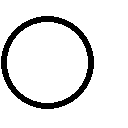}};
        \node[inner sep=0pt, below = 0.2cm of shape-circle-1, myshadow, label={[label distance=0.25cm]0: \Large ...}] (shape-circle-3){\includegraphics[width=0.75cm, height=0.75cm]{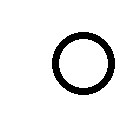}};

        \draw[dashed, very thick, draw=light-purple] (danger-sign-de-node) |- (danger-sign-image-box.east);
        \draw[dashed, very thick, draw=light-purple] (danger-sign-cn-node) |- (danger-sign-image-box_cn.west);
        \draw[dashed, very thick, draw=light-purple] (icon-animal-node) -- (icon-animal-node -| icon-animal-image-box.west);
        \draw[dashed, very thick, draw=light-purple] (shape-circle-node) -- (shape-circle-node -| shape-circle-image-box.west);

        \node (legend1) [below = 0.5cm of icon-animal-image-box, xshift=0.225cm]{\footnotesize \textbf{image instances}};
        \node (legend2) [below = 0.8cm of icon-animal-image-box, xshift=0cm]{\footnotesize \textbf{subClassOf}};

        \node (phase2) [rectangle, rounded corners, fill=light-yellow!40, minimum height=7cm, minimum width=24cm, below = 0.2cm of phase1, label={[label distance=-0.75cm,font=\bfseries, xshift=-10cm, yshift=-0.25cm]90: \Large 2. Training Phase}] {};

        \node[inner sep=0pt, myshadow2, left of=phase2, xshift=-8.25cm, yshift=0.5cm,] (danger-sign-t){\includegraphics[width=2cm, height=2cm]{images/RS_Danger1.png}};
        \node (image-embedding-legend) [draw=light-blue, fill=light-blue!30, minimum height=0.75cm, minimum width=0.75cm, below = 0.75cm of danger-sign-t, label={[label distance=0.1cm]0: \Large Image Emb.}]{};

        \node (node-embedding-legend) [draw=light-orange, fill=light-orange!30, minimum height=0.75cm, minimum width=0.75cm, right = 3cm of image-embedding-legend, label={[label distance=0.1cm]0: \Large Node Emb.}]{};
        \node (relation-embedding-legend) [draw=light-purple, fill=light-purple!30, minimum height=0.75cm, minimum width=0.75cm, below = 0.3cm of image-embedding-legend, label={[label distance=0.1cm]0: \Large Relation Emb.}]{};
        \node (triple-embedding-legend) [draw=light-green, fill=light-green!30, minimum height=0.75cm, minimum width=0.75cm, right = 3cm of relation-embedding-legend, label={[label distance=0.1cm]0: \Large Triplet}]{};
        \node (image-encoder) [encoder,  right = 1.5cm of danger-sign-t, text width=1.55cm, align=center] { \Large Image Encoder};
        
        \node (image-embedding-1) [draw=light-blue, fill=light-blue!30, minimum height=1.25cm, minimum width=1.25cm, right = 1.75cm of image-encoder]{\large $z^I$};

        \node (node-embedding-1) [draw=light-orange, fill=light-orange!30, minimum height=1.25cm, minimum width=1.875cm, yshift=1.5cm, right = 0.25cm of image-embedding-1]{\large $z^o_1$};
        \node (node-embedding-2) [draw=light-orange, fill=light-orange!30, minimum height=1.25cm, minimum width=1.875cm, right = 0cm of node-embedding-1]{\large $z^o_2$};
        \node (node-embedding-3) [draw=light-orange, fill=light-orange!30, minimum height=1.25cm, minimum width=1.875cm, right = 0cm of node-embedding-2]{\large $z^o_3$};
        \node (node-embedding-4) [draw=light-orange, fill=light-orange!30, minimum height=1.25cm, minimum width=1.875cm, right = 0cm of node-embedding-3]{\large ...};
        \node (node-embedding-5) [draw=light-orange, fill=light-orange!30, minimum height=1.25cm, minimum width=1.875cm, right = 0cm of node-embedding-4]{\large $z^o_{N_{o}}$};
        \node (triplet-1-4-1) [draw=light-green, fill=light-green!30, minimum height=1.25cm, minimum width=1.875cm, right = 0.25cm of image-embedding-1, xshift=0cm, yshift=-0cm]{};
        \node (triplet-1-4-2) [draw=light-green, fill=white, minimum height=1.25cm, minimum width=1.875cm, right = 0cm of triplet-1-4-1]{};
        \node (triplet-1-4-3) [draw=light-green, fill=white, minimum height=1.25cm, minimum width=1.875cm, right = 0cm of triplet-1-4-2]{};
        \node (triplet-1-4-4) [draw=light-green, fill=light-green!30, minimum height=1.25cm, minimum width=1.875cm, right = 0cm of triplet-1-4-3]{};
        \node (triplet-1-4-5) [draw=light-green, fill=white, minimum height=1.25cm, minimum width=1.875cm, right = 0cm of triplet-1-4-4]{};

        \node (triplet-1-3-1) [draw=light-green, fill=white, minimum height=1.25cm, minimum width=1.875cm, right = 0.25cm of image-embedding-1, xshift=0.15cm, yshift=-0.15cm]{};
        \node (triplet-1-3-2) [draw=light-green, fill=white, minimum height=1.25cm, minimum width=1.875cm, right = 0cm of triplet-1-3-1]{};
        \node (triplet-1-3-3) [draw=light-green, fill=light-green!30, minimum height=1.25cm, minimum width=1.875cm, right = 0cm of triplet-1-3-2]{};
        \node (triplet-1-3-4) [draw=light-green, fill=white, minimum height=1.25cm, minimum width=1.875cm, right = 0cm of triplet-1-3-3]{};
        \node (triplet-1-3-5) [draw=light-green, fill=white, minimum height=1.25cm, minimum width=1.875cm, right = 0cm of triplet-1-3-4]{};

        \node (triplet-1-2-1) [draw=light-green, fill=white, minimum height=1.25cm, minimum width=1.875cm, right = 0.25cm of image-embedding-1, xshift=0.3cm, yshift=-0.3cm]{};
        \node (triplet-1-2-2) [draw=light-green, fill=white, minimum height=1.25cm, minimum width=1.875cm, right = 0cm of triplet-1-2-1]{};
        \node (triplet-1-2-3) [draw=light-green, fill=white, minimum height=1.25cm, minimum width=1.875cm, right = 0cm of triplet-1-2-2]{};
        \node (triplet-1-2-4) [draw=light-green, fill=white, minimum height=1.25cm, minimum width=1.875cm, right = 0cm of triplet-1-2-3]{};
        \node (triplet-1-2-5) [draw=light-green, fill=light-green!30, minimum height=1.25cm, minimum width=1.875cm, right = 0cm of triplet-1-2-4]{};

        \node (triplet-1-1-1) [draw=light-green, fill=white, minimum height=1.25cm, minimum width=1.875cm, right = 0.25cm of image-embedding-1, xshift=0.45cm, yshift=-0.45cm]{};
        \node (triplet-1-1-2) [draw=light-green, fill=light-green!30, minimum height=1.25cm, minimum width=1.875cm, right = 0cm of triplet-1-1-1]{};
        \node (triplet-1-1-3) [draw=light-green, fill=white, minimum height=1.25cm, minimum width=1.875cm, right = 0cm of triplet-1-1-2]{};
        \node (triplet-1-1-4) [draw=light-green, fill=white, minimum height=1.25cm, minimum width=1.875cm, right = 0cm of triplet-1-1-3]{};
        \node (triplet-1-1-5) [draw=light-green, fill=white, minimum height=1.25cm, minimum width=1.875cm, right = 0cm of triplet-1-1-4]{};

        \node (triplet-1-0-1) [draw=light-green, fill=light-green!30, minimum height=1.25cm, minimum width=1.875cm, right = 0.25cm of image-embedding-1, xshift=0.6cm, yshift=-0.6cm]{ $(z^I,z^r_0,z^o_1)$};
        \node (triplet-1-0-2) [draw=light-green, fill=white, minimum height=1.25cm, minimum width=1.875cm, right = 0cm of triplet-1-0-1]{  $(z^I,z^r_0,z^o_2)$};
        \node (triplet-1-0-3) [draw=light-green, fill=white, minimum height=1.25cm, minimum width=1.875cm, right = 0cm of triplet-1-0-2]{ $(z^I,z^r_0,z^o_3)$};
        \node (triplet-1-0-4) [draw=light-green, fill=white, minimum height=1.25cm, minimum width=1.875cm, right = 0cm of triplet-1-0-3]{\large ...};
        \node (triplet-1-0-5) [draw=light-green, fill=light-green!30, minimum height=1.25cm, minimum width=1.875cm, right = 0cm of triplet-1-0-4]{ $(z^I,z^r_0,z^o_{N_o})$};

        \node (rel-embedding-4) [draw=light-purple, fill=light-purple!30, minimum height=1.25cm, minimum width=1.25cm, right = 0.75cm of node-embedding-5, yshift=-0.25cm]{\large $z_{N_{r}}^r$};
        \node (rel-embedding-3) [draw=light-purple, fill=light-purple!30, minimum height=1.25cm, minimum width=1.25cm, right = 0.75cm of node-embedding-5, yshift=-0.4cm, xshift=0.15cm]{\large $z_{N_{r}}^r$};
        \node (rel-embedding-2) [draw=light-purple, fill=light-purple!30, minimum height=1.25cm, minimum width=1.25cm, right = 0.75cm of node-embedding-5, yshift=-0.55cm,xshift=0.3cm]{\large $z_{N_{r}}^r$};
        \node (rel-embedding-1) [draw=light-purple, fill=light-purple!30, minimum height=1.25cm, minimum width=1.25cm, right = 0.75cm of node-embedding-5, yshift=-0.7cm, xshift=0.45cm]{\large $z_1^r$};
        \node (rel-embedding-0) [draw=light-purple, fill=light-purple!30, dashed, thick, minimum height=1.25cm, minimum width=1.25cm, right = 0.75cm of node-embedding-5, yshift=-0.85cm, xshift=0.6cm]{\large $z_0^r$};
        
        \draw [decorate,decoration={brace, amplitude=1mm, raise=.5mm}] (triplet-1-4-5.north east) -- (triplet-1-1-5.north east) node[above right, midway] { $N_{r}$};
        \draw [decorate,decoration={brace, amplitude=1mm, raise=.5mm}] (rel-embedding-4.north east) -- (rel-embedding-1.north east) node[above right, midway] { $N_{r}$};

        \node (decoder) [encoder,  below = 0.5cm of triplet-1-0-1, text width=1.55cm, align=center, xshift=-0.5cm] { \Large Decoder};
        \node (cross-entropy) [rectangle, right=1cm of decoder, text width=2cm, draw = black, thick, minimum height=1.5cm, minimum width=2.5cm, align=center, fill=white]{\Large Cross Entropy Loss};
        \node (y_2) [rectangle, right=1cm of cross-entropy, text width=0.5cm, draw = black, thick, minimum height=2cm, minimum width=1cm, align=center, xshift=0.15cm, yshift=0cm,fill=white]{\Large $y$};
        \node (regularization-loss) [rectangle, right=1.5cm of y_2, text width=2cm, draw = black, thick, minimum height=1.5cm, minimum width=2.5cm, align=center, fill=white]{\Large Regular. Loss};

        \draw[dashed, very thick, draw=light-purple] (legend1.west) -- +(-0.8cm, 0cm);
        \draw[arrow, dashed, very thick, draw=light-purple] (legend2.west) -- +(-0.8cm, 0cm);

        \draw[arrow] (danger-sign-t.east) -- (image-encoder.west);
        \draw[arrow] (image-encoder.east) -- +(1cm,0cm) |-(image-embedding-1.west);
        \draw[arrow] (image-embedding-1.south) |-(decoder.west);
        \draw[arrow] (decoder.east) -- (cross-entropy);
        \draw[arrow] (y_2.west) -- (cross-entropy.east);
        \draw[arrow] (triplet-1-0-5.east)  -| (regularization-loss.north);
        \draw[arrow] (knowledge-graph.south -| node-embedding-3.north) -- node [right,midway, yshift=-0.3cm] {Node Embeddings} (node-embedding-3.north);
        \draw[arrow] (knowledge-graph.south -| rel-embedding-2.north) -- node [right,midway, text width=2cm, yshift=0cm] {Relation\\Embeddings} (rel-embedding-2.north);
        \node (phase3) [rectangle, rounded corners, fill=light-green!10, minimum height=15cm, minimum width=5cm, right = 0.2cm of phase2, yshift=4cm, label={[label distance=-0.75cm,font=\bfseries ]90: \Large 3. Inference Phase}] {};
        \node[inner sep=0pt, above of = phase3, myshadow2, yshift=3.5cm] (rs-danger-i){\includegraphics[width=2cm, height=2cm]{images/RS_Danger.png}};
        \node (image-encoder) [encoder,  below = 1.25cm of rs-danger-i, text width=1.55cm, align=center] { \Large Image Encoder};
        \node (node-embedding-1) [draw=light-blue, fill=light-blue!30, minimum height=1.25cm, minimum width=1.25cm, below = 1.5cm of image-encoder]{\large $z^I$};
        \node (decoder) [encoder,  below = 1.25cm of node-embedding-1, text width=1.55cm, align=center] { \Large Decoder};
        \node (y_2) [rectangle, below=1.25cm of decoder, text width=0.5cm, draw = black, thick, minimum height=1cm, minimum width=1cm, align=center, fill=white, xshift=-0.5cm, fill=gray!20]{\Large $\hat{y}_2$};
        \node (y_1) [rectangle, left=0cm of y_2, text width=0.5cm, draw = black, thick, minimum height=1cm, minimum width=1cm, align=center]{\Large $\hat{y}_1$};
        \node (y_3) [rectangle, right=0cm of y_2, text width=0.5cm, draw = black, thick, minimum height=1cm, minimum width=1cm, align=center, fill=white]{\Large ...};
        \node (y_4) [rectangle, right=0cm of y_3, text width=0.5cm, draw = black, thick, minimum height=1cm, minimum width=1cm, align=center, fill=white]{\Large $\hat{y}_{N_{\textrm{cls}}}$};
        \draw[arrow] (rs-danger-i.south) -- (image-encoder.north);
        \draw[arrow] (image-encoder.south) -- (node-embedding-1.north);
        \draw[arrow] (node-embedding-1.south) -- (decoder.north);
       \draw[arrow] (decoder.south) -- (decoder.north|- y_2.north);
    \end{tikzpicture}
    \end{adjustbox}
    \caption{\small The KGV architecture --
    Our approach consists of three phases, namely knowledge modeling, training, and inference.
    \textbf{Knowledge modeling phase:} We create a knowledge graph based on prior domain knowledge. Also, synthetic images are generated for object categories (e.g., shapes and colors) that are semantically represented in the knowledge graph but lack visual information in the dataset.
    \textbf{Training phase:} The neural network is fed with both synthetic and real-world images and trained end-to-end by adding the regularization loss and cross-entropy loss together as a total loss for optimization.
    The image embeddings $\boldsymbol{z^I}$ and knowledge graph embeddings $\boldsymbol{z^r_i}$, $\boldsymbol{z^o_j}$ are aligned by minimizing the regularization loss. The Cross-entropy loss is used to classify images based on their image embedding representation.
    \textbf{Inference phase:} The classification task is completed by selecting the class with the highest possibility based on the output of the decoder.}
    \label{fig:architeture}
\vspace{-1em}
\end{figure*}

%% file: sections/04_experiment.tex
\section{Experiments}
\label{sec:experiment}
In this section, we first explain the datasets and the respective knowledge graphs used in our experiments. 
Then, we analyze the results of supervised image classification experiments under varying conditions, including data distribution shifts, low-data regimes, and few-shot learning scenarios. 
The experimental details are listed in the appendix section. 

\subsection{Datasets}
In our experiments, we evaluate KGV in three domains, namely Road Sign, Car Recognition, and ImageNet. 
\begin{itemize}

    \item \textbf{Road Sign Recognition:} It includes three commonly used road sign datasets from Germany, China, and Russia: German Traffic Sign Recognition Benchmark (GTSRB) from \citet{6033395}, Chinese Traffic Sign Dataset (CTSD) from \citet{yang2015towards}, and Russian Traffic Sign Images Dataset (RTSD) from \citet{shakhuro2016russian}. 
    \item \textbf{Car Recognition:} We use the Deep Visual Marketing Car (DVM-CAR) dataset ~\cite{DBLP:journals/corr/abs-2109-00881} which contains \emph{millions} of car images captured under various conditions and viewpoints, accompanied by detailed annotations including car make, seat count, door count, color, body type and etc. The primary task here is to classify the car model based on the provided image.
    \item \textbf{ImageNet:} In this domain, we use the well-known mini-ImageNet~\cite{vinyals2016matching}, ImageNet-Sketch~\cite{wang2019learning}, ImageNetV2~\cite{recht2019imagenet}, ImageNet-R~\cite{hendrycks2021many}, and ImageNet-A~\cite{hendrycks2021natural} for our experiments. 

\end{itemize}

\subsection{Knowledge Graphs}
\subsubsection{Road Sign Recognition Knowledge Graph:}
The constructed knowledge graph contains 202 nodes and 5 relations. 
The 5 relations we include are `has the background color of', `has the border color of', `has the sign legend of', `has the shape of', `instance/subclass'. 

\subsubsection{Car Recognition Knowledge Graph:}
The knowledge graph for the DVM-CAR contains 386 nodes and 9 relations. 
These relations are `hasBodyType', `hasColor', `hasDoorNo', `hasFuelType', `hasGearbox', `hasMaker', `hasSeatNo', `hasViewpoint', `instanceOf'. 
The dataset includes annotations such as body type, color, and car make for each image, provided in tabular form. 
We construct a KG by defining the ontology and triplets based on the tabular annotations. 
We generate pure color images using the same method as in the road sign domain to serve as visual prior knowledge. 
Further, we collect sketch images representing body types such as SUV, Coupe, and Limousine.
Examples of these body type sketches are provided in the appendix.

\subsubsection{ImageNet Knowledge Graph:}
We leverage WordNet \cite{miller1995wordnet, fellbaum2010wordnet}, which is a lexical database containing nouns, verbs, adjectives, and adverbs of the English language structured into respective synsets and create the subset of Wordnet by extracting the respective synsets of each label from the mini-ImageNet dataset. These synsets are grouped based on the lexical domain they pertain to, e.g., animal, artifact, or food; thus, we get hierarchical relations between the classes in the mini-ImageNet from the WordNet. We also include two association relations: `has color' and `has sketch'. We use the generated pure color images in the domain of road signs and the sketch images from ImageNet-Sketch. It contains 234 nodes and 3 relations: `subclass/instance of', `has color' and `has sketch'.


\subsection{Results}
We conducted a number of supervised image classification experiments with major and minor distribution shifts and low data regimes. 
Further, we evaluated the model's adaptability on few-shot learning experiments using data with altered distributions.
\subsubsection{Image Classification under Data Distribution Shifts:}
\begin{table}[ht]
    \centering
    \resizebox{.48\textwidth}{!}{
    \begin{tabular}{c c c c c c c}
        \toprule
        \multicolumn{3}{c}{Prior Know.} & \multirow{2}{*}{Model} & \multirow{2}{*}{GTSRB (43)} & \multirow{2}{*}{CTSD (25)} & \multirow{2}{*}{RTSD (36)} \\
        Imp. & KG & VP & & & &  \\
        \midrule
        \xmark & \xmark & \xmark & ResNet50 & 98.9\% \textcolor{light-green}{(+0.4\%)} & 71.5\% \textcolor{light-green}{(+3.3\%)} & 72.8\% \textcolor{light-green}{(+7.3\%)} \\
        \cmark & \xmark & \xmark & CLIP+LP & 77.2\% \textcolor{light-green}{(+22.1\%)} & 73.5\% \textcolor{light-green}{(+1.3\%)}& 60.0\%\textcolor{light-green}{(+20.1\%)}\\
        \cmark & \xmark & \xmark & DINOv2+LP & 78.7\% \textcolor{light-green}{(+20.6\%)} & 28.9\% \textcolor{light-green}{(+49.5\%)}& 59.9\% \textcolor{light-green}{(+20.2\%)}\\
        \xmark & \cmark & \cmark  & DGP & 99.0 \%\textcolor{light-green}{(+0.3\%)}  & 72.0\% \textcolor{light-green}{(+2.8\%)}& 73.5\% \textcolor{light-green}{(+6.6\%)}\\
        \xmark & \cmark & \cmark  & GCNZ & 99.1\% \textcolor{light-green}{(+0.2\%)}& 71.1\% \textcolor{light-green}{(+3.7\%)}& 74.9\% \textcolor{light-green}{(+5.2\%)}\\
        \xmark & \cmark & \cmark  & KGV (ours) & \textbf{99.3}\% & \textbf{74.8}\% & \textbf{80.1}\%\\
        
        \midrule
        \xmark & \xmark & \cmark  & ResNet50$^+$ & 98.9\% & 71.8\% & 72.7\%\\
        \xmark & \cmark & \xmark  & KGV$^-$ & 99.0\% & 72.5\% & 74.0\%\\
        \bottomrule
    \end{tabular}
    }

    \caption{Road Sign Classification Accuracy. Evaluation results with the test data of GTSRB and the full data of CTSD and RTSD.
    Imp. refers to the implicit prior knowledge in pre-trained models, while KG indicates explicit prior knowledge from the knowledge graph. VP denotes the visual prior associated with object category elements.
    }
    \label{tab:overall results}
\end{table}

\begin{table}[ht]
    \centering
    \resizebox{.37\textwidth}{!}{
    \begin{tabular}{c c c c c c }
        \toprule
        \multirow{2}{*}{Classes} & \multirow{2}{*}{Images} & \multicolumn{2}{c}{ResNet50} & \multicolumn{2}{c}{KGV} \\

          &  & Precision & Recall & Precision & Recall  \\
        \midrule
          STOP & \makecell{\includegraphics[width=0.6cm,height=0.6cm]{images/RS_Stop.png} \includegraphics[width=0.6cm,height=0.6cm]{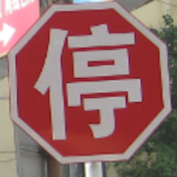}} & 21.3\% & 74.6\% & \textbf{48.2}\% & \textbf{98.1}\%\\
        \midrule
         Danger & \makecell{\includegraphics[width=0.6cm,height=0.6cm]{images/RS_Danger.png} \includegraphics[width=0.6cm,height=0.6cm]{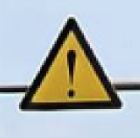}} & 0.0\% & 0.0\% & \textbf{10.8}\% & \textbf{14.9}\% \\
        \midrule
        BendLeft & \makecell{\includegraphics[width=0.6cm,height=0.6cm]{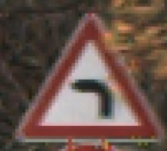} \includegraphics[width=0.6cm,height=0.6cm]{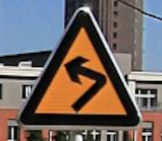}} & 0.0\% & 0.0\% & \textbf{3.5}\% & \textbf{32.1}\%\\

        \midrule
        PedestrianCross & \makecell{\includegraphics[width=0.6cm,height=0.6cm]{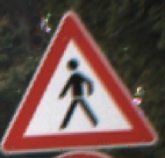} \includegraphics[width=0.6cm,height=0.6cm]{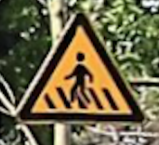}} & 0.0\% & 0.0\% & \textbf{9.1} \% & \textbf{65.6}\%  \\
        \midrule
         SchoolCross & \makecell{\includegraphics[width=0.6cm,height=0.6cm]{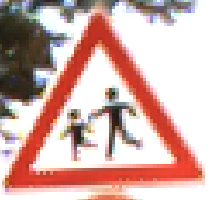} \includegraphics[width=0.6cm,height=0.6cm]{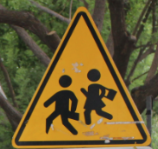}} & 0.0\% & 0.0\% & \textbf{20.1}\% & \textbf{25.4}\% \\
        \bottomrule
    \end{tabular}
    }
    \caption{Precision and recall of representative classes for road sign classification when training with 43 classes data of GTSRB and testing with shared 25 classes data of CTSD.}
    \label{tab:precision and recall results}
\end{table}
We train our KGV model and ResNet50 with the GTSRB dataset plus generated synthetic images from scratch.
It is then evaluated with test data of GTSRB with 43 classes, the full data of CTSD with 25 shared classes, and the full data of RTSD with 36 shared classes. 
The shared classes here indicate those classes that both datasets contain. 
As shown in Table \ref{tab:overall results}, all models' accuracy drops in distribution shift scenarios, i.e., from Germany to China or Russia. 
By incorporating the multi-modal prior knowledge, our KGV reaches an obvious improvement compared to other baselines, such as DGP \cite{Kampffmeyer_2019_CVPR} and GCNZ \cite{wang2018zero}. These two are ZSL baselines, which leverage additional information from KGs. 
Fine-tuning pretrained CLIP and DINOv2 with linear probing does not work well in road sign domain. 
`ResNet50$^+$' refers to the model trained with additional synthetic images and `KGV$^-$' denotes the model trained without synthetic images. 
As the experiments indicate, our KGV model surpasses the ResNet50 baseline by 3.3\% in CTSD and 7.3\% in RTSD.
We also observe that the inclusion of synthetic images featuring elements of object categories, such as colors and shapes, enhances the generalization ability of the KGV models across various data distributions. 
On the other hand, it does not have a significant effect on the ResNet50's performance. 
The precision and recall for a subset of classes in CTSD are presented in Table \ref{tab:precision and recall results}. 
As the table illustrates, the ResNet50 fails to classify specific classes, including danger signs, pedestrian crossing, etc. 
In contrast, the KGV models exhibit significant improvements in both precision and recall.

\begin{table}[ht]
\centering
\resizebox{.48\textwidth}{!}{
    \begin{tabular}{ c c c c c c | c c}
    \toprule
    Datasets & Samples & ResNet50 & KGNN & ResNet50$^+$ & KGV & CLIP+LP & DINOv2+LP \\ 
    \midrule
    \small Mini-ImageNet & \makecell{\includegraphics[width=0.7cm, height=0.7cm]{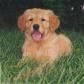} \includegraphics[width=0.7cm, height=0.7cm]{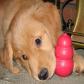} \includegraphics[width=0.7cm, height=0.7cm]{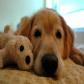}} & \makecell{62.6\% \\ \textcolor{light-green}{+4.2\%}} & \makecell{66.2\% \\ \textcolor{light-green}{+0.6\%}} & \makecell{59.6\% \\ \textcolor{light-green}{+7.2\%}} & \textbf{66.8}\% & 87.0\% & 94.9\%\\ 
    ImageNetV2 & \makecell{\includegraphics[width=0.7cm, height=0.7cm]{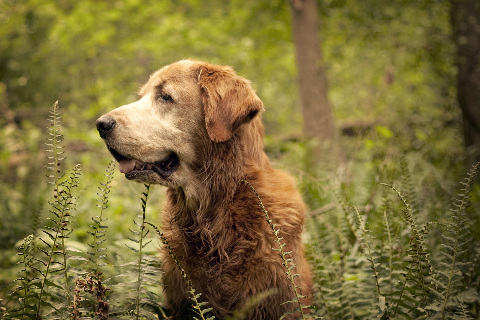} \includegraphics[width=0.7cm, height=0.7cm]{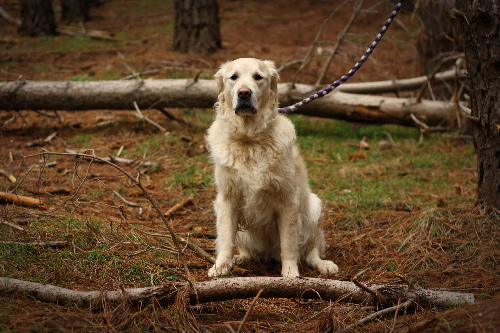} \includegraphics[width=0.7cm, height=0.7cm]{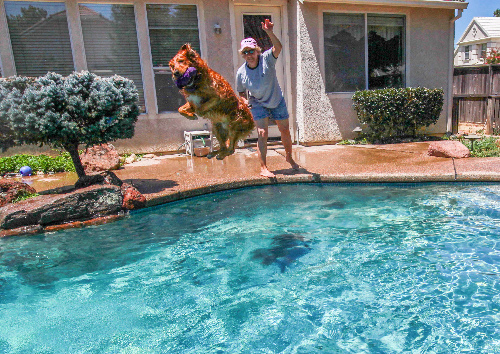}} & \makecell{50.5\% \\ \textcolor{light-green}{+2.2\%}} & \makecell{52.5\% \\ \textcolor{light-green}{+0.2\%}} & \makecell{48.0\% \\ \textcolor{light-green}{+4.7\%}} & \textbf{52.7}\% & 87.5\% & 91.5\%\\ 
    ImageNet-R & \makecell{\includegraphics[width=0.7cm, height=0.7cm]{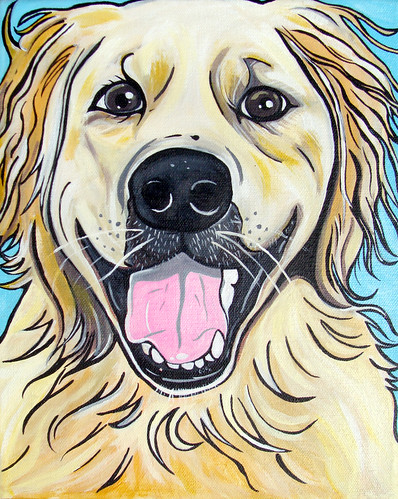} \includegraphics[width=0.7cm, height=0.7cm]{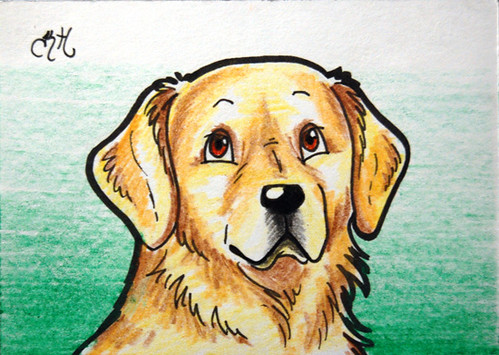} \includegraphics[width=0.7cm, height=0.7cm]{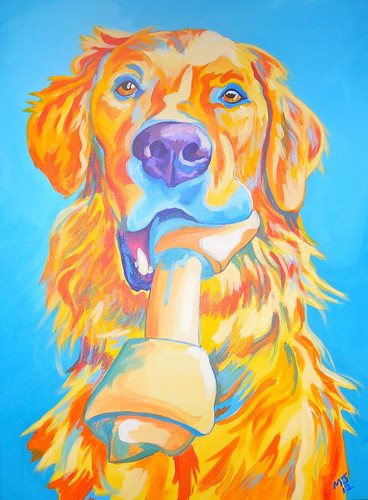}}  & \makecell{25.5\% \\ \textcolor{light-green}{+15.9\%}} & \makecell{30.0\% \\ \textcolor{light-green}{+11.4\%}} & \makecell{36.9\% \\ \textcolor{light-green}{+4.5\%}} & \textbf{41.4}\% & 90.6\% & 86.1\%\\ 
    ImageNet-A & \makecell{\includegraphics[width=0.7cm, height=0.7cm]{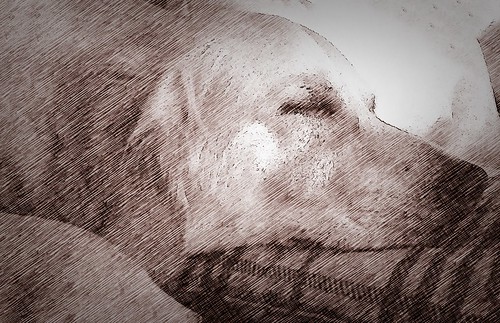} \includegraphics[width=0.7cm, height=0.7cm]{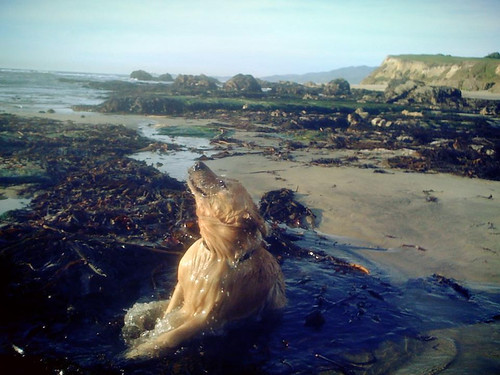} \includegraphics[width=0.7cm, height=0.7cm]{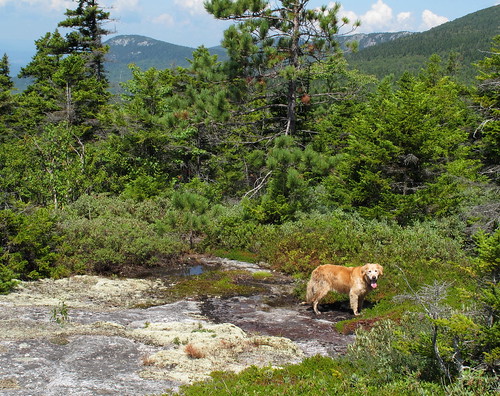}} & \makecell{4.7\% \\ \textcolor{light-green}{+2.1\%}} & \makecell{5.3\% \\ \textcolor{light-green}{+1.5\%}} & \makecell{5.0\% \\ \textcolor{light-green}{+1.8\%}} & \textbf{6.8}\% & 53.8\%& 84.5\%\\ 
    \bottomrule
    \end{tabular}
}
\caption{Classification Accuracy in the ImageNet Domain. ResNet50$^+$ refers to the ResNet50 model trained with additional visual prior images.}
\label{tab: imagenet}
\end{table}
We also evaluate our method in the ImageNet domain. The baselines we choose are ResNet50, DINOv2, CLIP, and KGNN~\cite{DBLP:conf/semweb/MonkaHR22}. KGNN represents another approach to integrate prior knowledge into learning pipeline through pre-training. 
Our model outperforms the baselines KGNN and ResNet50 in all these datasets.
As Table \ref{tab: imagenet} shows, these baselines fail, particularly when the distribution shift increases (from Mini-ImageNet to ImageNetV2, ImageNet-R, and ImageNet-A). 
Our KGV suffers less when data distribution changes. 
We also observe that incorporating visual priors without the guidance of symbolic knowledge from the KG may negatively impact performance in scenarios where the difference between training and test distributions is minimal. Compared to ResNet50, ResNet50$^{+}$’s performance decreases from 62.6\% to 59.6\% on mini-ImageNet and from 50.5\% to 48.0\% on ImageNetV2. We believe this is due to the visual priors altering the distribution of the training data, making it different from the test distribution.
However, DINOv2 and CLIP fine-tuned with linear probing achieve better performance than our model, as they are pre-trained on massive data, including ImageNet and its variants. 
We conducted additional experiments by combining DINOv2 and CLIP with our approach, respectively, as described in Section~\ref{sec:implicit and explicit}.

\begin{table}[ht]
\centering
\resizebox{.48\textwidth}{!}{
        \begin{tabular}{c | c c c c c c c}
        \toprule
            Model & SimCLR & MMCL & SupCon & ResNet50 & CLIP+LP & DINOv2+LP & KGV (Ours) \\
        \midrule
            ACC & \makecell{88.8\% \\ \textcolor{light-green}{+7.5\%}} & \makecell{94.4\% \\ \textcolor{light-green}{+1.9\%}} & \makecell{93.8\% \\ \textcolor{light-green}{+2.5\%}} & \makecell{94.3\% \\ \textcolor{light-green}{+2.0\%}} & \makecell{65.8\% \\ \textcolor{light-green}{+30.5\%}} & \makecell{89.3\% \\ \textcolor{light-green}{+7.0\%}} &  \textbf{96.3}\% \\
        \bottomrule
        \end{tabular}
}
\caption{Accuracy in DVM-CAR Classification.}
\label{tab: DVM-CAR dataset}
\end{table}

Next, we evaluate KGV in the DVM-CAR dataset.
We observe that even though the test and training data have no significant distribution shift, it still improves performance.
The baselines we choose are contrastive learning approaches such as SimCLR~\cite{pmlr-v119-chen20j} and SupCon~\cite{DBLP:journals/corr/abs-2004-11362} as well as the SOTA model MMCL~\cite{Hager_2023_CVPR}, which combines the tabular data and image data to enhance the model's performance through contrastive learning. 
Table~\ref{tab: DVM-CAR dataset} shows that our model outperforms all baselines.
In the mini-ImageNet domain, KGV achieves a 4.2\% performance gain compared to ResNet50. 

\subsubsection{Performance in Low Data Regimes:}
\begin{figure}[ht]
    \centering
    \subfigure[Road Sign Classification in GTSRB.]{
        \includegraphics[width=0.22\textwidth]{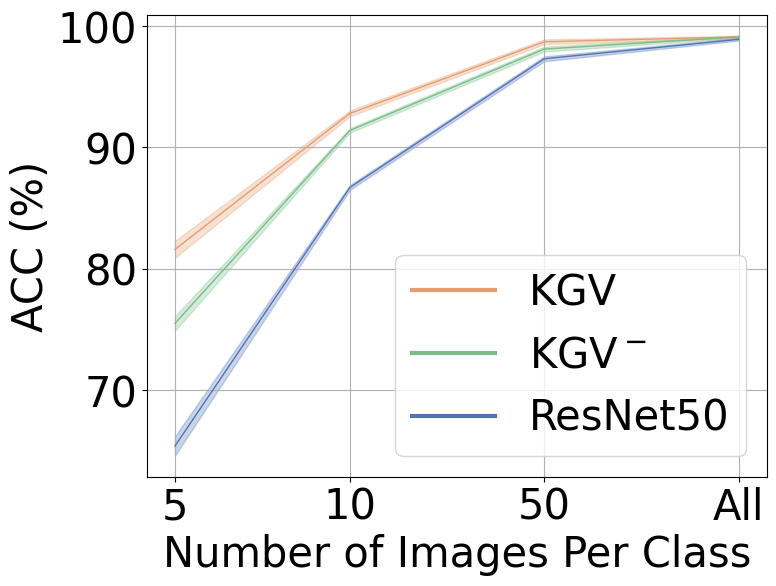}
        \label{fig:subfig1}
    }
    \hfill
    \subfigure[DVM-CAR Model Classification.]{
        \includegraphics[width=0.22\textwidth]{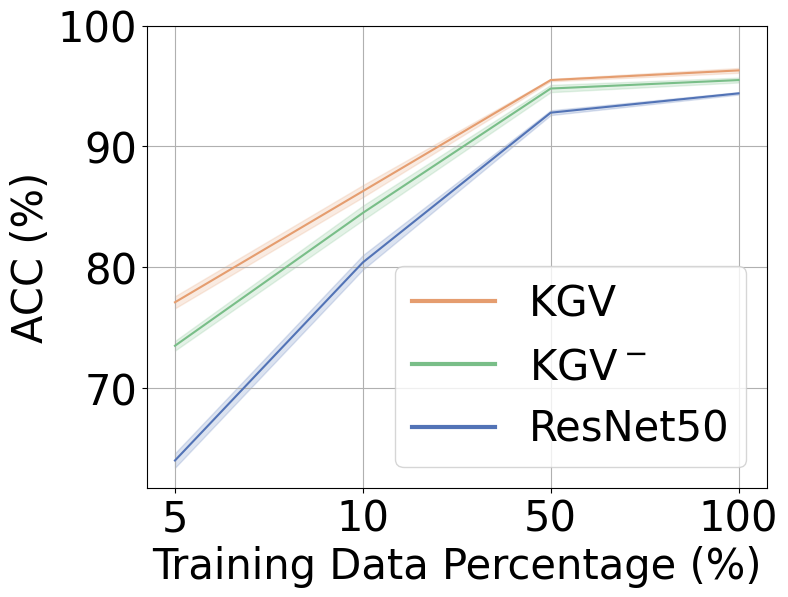}
        \label{fig:subfig2}
    }
    \caption{Performance under Low Data Regime. KGV$^-$ is the KGV variant trained without synthetic images.}
    \label{fig:low data regimes}
\end{figure}

Figure~\ref{fig:low data regimes} 
shows the evaluation results KGV in low data regimes. 
We see that our method significantly outperforms the ResNet50 baseline, particularly when the amount of training data is limited. 
With the increase of training data, the performance of the two models gradually converges. 
However, KGV consistently outperforms the baseline among all training settings. 
The performance of KGV$^-$ -- the variant of KGV trained without synthetic images -- 
is also shown in Figure~\ref{fig:low data regimes}. 
It indicates that without the visual priors, its performance drops. 

\subsubsection{Few-shot Learning:}
\begin{table}[ht]
    \resizebox{.42\textwidth}{!}{
        \centering
        \begin{tabular}{ c c l l}
        \toprule
        Experiments & Models & CTSD (58 classes) & RTSD (64 classes) \\ 
        \midrule
        \multirow{2}{*}{One-shot} & ResNet50 & 68.4\% \textcolor{light-green}{(+7.0\%)} & 57.9\% \textcolor{light-green}{(+5.2\%)}\\ 
         & KGV & \textbf{75.4}\%  & \textbf{63.1}\% \\ 
        \midrule
        \multirow{2}{*}{Five-shot} & ResNet50 & 98.7\% \textcolor{light-green}{(+0.2\%)} & 81.2\% \textcolor{light-green}{(+2.3\%)}\\ 
         & KGV & \textbf{98.9}\%  & \textbf{83.5}\% \\ 
        \bottomrule
        \end{tabular}
    }
    \caption{One-shot and five-shot learning results in the datasets of CTSD and RTSD are demonstrated in the table.}
    \label{tab:few-shot learning}
\end{table}
\begin{table*}[!h]
\centering
\resizebox{.97\textwidth}{!}{
    \begin{tabular}{c c | c | c c c | c c c c | c}
    \toprule        
    \multicolumn{2}{c|}{Prior Knowledge} & \multirow{2}{*}{Model} & \multicolumn{3}{c|}{Road Sign Domain} & \multicolumn{4}{c|}{ImageNet Domain} & Car Domain \\
        Imp. & KG & & GTSRB & CTSD & RTSD & Mini-ImageNet & ImageNetV2 & ImageNet-R &  ImageNet-A & DVM-CAR  \\
    \midrule    
    \cmark & \cmark & P-ResNet50+KGV & 99.5\% & 74.5\% & 79.8\% & 81.8\% & 62.4\% & 56.6\% & 6.4\% & 97.0 \%\\ 
    \cmark & \xmark & P-ResNet50 & 99.3\% \textcolor{light-green}{(+0.2\%)} & 73.1\% \textcolor{light-green}{(+1.4\%)}& 72.5\% \textcolor{light-green}{(+7.3\%)} & 75.0\%  \textcolor{light-green}{(+6.8\%)} & 61.3\%  \textcolor{light-green}{(+1.1\%)}& 51.6\%  \textcolor{light-green}{(+5.0\%)} & 4.4\%  \textcolor{light-green}{(+2.0\%)}& 96.2\%  \textcolor{light-green}{(+0.8\%)}\\ 
    \midrule
    \cmark & \cmark & DINOv2+KGV & 79.1\% & 31.2\% & 62.4\% & 95.4\% & 92.4\% & 87.0\% & 85.4\% & 92.8\% \\
    \cmark & \xmark & DINOV2+LP & 78.7\% \textcolor{light-green}{(+0.4\%)}& 28.9\%  \textcolor{light-green}{(+2.3\%)}& 59.9\%  \textcolor{light-green}{(+2.5\%)} & 94.9\%  \textcolor{light-green}{(+0.6\%)} & 91.5\%  \textcolor{light-green}{(+0.9\%)} & 86.1\%  \textcolor{light-green}{(+0.9\%)} & 84.3\%  \textcolor{light-green}{(+1.1\%)} & 89.3\%  \textcolor{light-green}{(+3.5\%)} \\ 
    \midrule
    \cmark & \cmark & CLIP-I+KGV & 89.3\% & 73.0\% & 67.8\% & 90.1\% & 89.1\% & 83.2\% & 50.1\% & 90.5\% \\ 
    \cmark & \xmark & CLIP-I+LP & 89.1\%  \textcolor{light-green}{(+0.2\%)} & 73.5\%  \textcolor{light-red}{(-0.5\%)} & 60.0\%  \textcolor{light-green}{(+7.8\%)}& 89.7\%  \textcolor{light-green}{(+0.4\%)}& 88.9\%  \textcolor{light-green}{(+0.2\%)} & 82.9\%  \textcolor{light-green}{(+0.3\%)} & 49.6\%  \textcolor{light-green}{(+0.5\%)}& 89.2\% \textcolor{light-green}{(+1.3\%)} \\ 
    \bottomrule
    \end{tabular}
}
\caption{Pre-trained ResNet50, DINOv2 (1B), and CLIP Image encoder, which are denoted as P-ResNet50, DINOv2, and CLIP-I, are used for our experiments. LP represents linear probing for fine-tuning. Our KGV's encoder is replaced by one of these pre-trained image encoders denoted by X+KGV.}
\label{tab: implicit_explicit prior}
\end{table*}

\begin{table*}[!h]
\centering
\resizebox{.97\textwidth}{!}{
    \begin{tabular}{c | c c c | c c c c | c}
    \toprule        
    \multirow{2}{*}{Model} & \multicolumn{3}{c|}{Road Sign Domain} & \multicolumn{4}{c|}{ImageNet Domain} & Car Domain \\
    & GTSRB & CTSD & RTSD & Mini-ImageNet & ImageNetV2 & ImageNet-R &  ImageNet-A & DVM-CAR  \\
    \midrule
    KGV & 99.3\% & \textbf{74.8\%} & \textbf{80.1\%} & \textbf{66.8\%} & \textbf{52.7\%} & \textbf{41.4\%} & \textbf{6.8\%} & \textbf{96.3\%}\\
    KGV w/o Gaussian Emb. &  \textbf{99.4\%} \textcolor{light-red}{(-0.1\%)}& 74.7\% \textcolor{light-green}{(+0.1\%)}& 79.9\% \textcolor{light-green}{(+0.2\%)}& 61.2\% \textcolor{light-green}{(+5.6\%)}& 49.0\% \textcolor{light-green}{(+3.7\%)} & 37.8\% \textcolor{light-green}{(+3.6\%)}& 4.8\% \textcolor{light-green}{(+2.0\%)} & 96.3\%\textcolor{light-green}{(+0.0\%)}\\
    KGV w/ TransH & 99.3\% \textcolor{light-green}{(0.0\%)}& 73.5\% \textcolor{light-green}{(+1.3\%)} & 79.3\% \textcolor{light-green}{(+0.0\%)}& 61.8\% \textcolor{light-green}{(+5.0\%)}& 48.2\% \textcolor{light-green}{(+4.5\%)} & 38.6\% \textcolor{light-green}{(+2.8\%)}& 5.5\% \textcolor{light-green}{(+1.3\%)} & 95.8\% \textcolor{light-green}{(+0.5\%)}\\
    KGV w/ DistMult & 99.2\% \textcolor{light-green}{(+0.1\%)} & 73.7\% \textcolor{light-green}{(+1.1\%)} & 75.6\% \textcolor{light-green}{(+5.4\%)}& 60.5\% \textcolor{light-green}{(+6.3\%)}	& 47.6\% \textcolor{light-green}{(+5.1\%)}& 37.5\% \textcolor{light-green}{(+3.1\%)}  & 4.7\% \textcolor{light-green}{(+2.1\%)} & 95.1\% \textcolor{light-green}{(+1.2\%)}\\
    KGV w/ RGCN & 99.3\% \textcolor{light-green}{(+0.0\%)} & 74.1\% \textcolor{light-green}{(+0.7\%)} & 78.5\% \textcolor{light-green}{(+1.6\%)} & 63.5\% \textcolor{light-green}{(+3.3\%)} & 50.6\% \textcolor{light-green}{(+2.1\%)} & 39.7\% \textcolor{light-green}{(+1.7\%)}& 6.1\% \textcolor{light-green}{(+0.7\%)}& - \\
    \bottomrule
    \end{tabular}
}
\caption{KGV with alternative score function designs like TransH, DistMult, and replacing Gaussian embeddings with vector embeddings are listed in the table. KGV with GNN-based method like RGCN are also compared.}
\label{tab: ablation kge}
\end{table*}
We conduct our few-shot learning tasks in the domain of road sign recognition. We first train the models with the GTSRB dataset. 
Then, we keep the encoder network and replace the decoder to fit the right class dimension in the target dataset. 
The new decoder is randomly initialized. 
The models are retrained with different amounts of labeled target data in the CTSD or the RTSD. 
We perform one-shot and five-shot learning experiments. One-shot learning involves using just one image per class from the target dataset for retraining. 
Similarly, five-shot learning uses five images per class in the target dataset for retraining. 
As Table~\ref{tab:few-shot learning} shows that KGV achieves 7.0\% performance gain in the CTSD dataset and 5.2\% in the RTSD dataset in one-shot learning. 
In a five-shot learning setting, the KGV model achieves 0.2 \% improvement in CTSD and 2.3\% improvement in RTSD. 

\subsubsection{Combining Implicit and Explicit Prior Knowledge}
\label{sec:implicit and explicit}
In this sub-section, we investigate if combining implicit prior knowledge from the pre-trained models and explicit multi-modal prior knowledge from KG and visual prior images can further improve our method's performance. 
We conduct experiments by replacing the random initialized ResNet50 encoder with pre-trained ResNet50, DINOv2(1B), and the image encoder of CLIP. 
As shown in Table~\ref{tab: implicit_explicit prior}, applying our method to these pre-trained encoders consistently enhances performance across all three domains. Although DINOv2 and CLIP have already achieved strong results on the ImageNet domain, the incorporation of KGV further improves their performance. 
These results indicate that our method can be adapted to state-of-the-art pre-trained image encoders to further enhance their performance.

\subsection{Ablation Study}
\subsubsection{Score Functions}
Here, we compare four score function designs: (1) replacing the Gaussian-form node embedding with vector embedding (TransE), (2) using TransH’s score function, (3) adopting DistMult’s score function, and (4) incorporating R-GCN, a GNN-based method. As Table~\ref{tab: ablation kge} shows, overall KGV performs best among all these score function designs. 
The performance of KGV without Gaussian embedding drops significantly in the ImageNet domain compared to other domains. 
This highlights the importance of using Gaussian embedding for node representation to handle inclusion relations between object categories and their respective images. 
The DistMult demonstrates suboptimal performance across all three visual domains. While GNN-based method R-GCN outperform TransH and DistMult, it still does not surpass the effectiveness of our original score function design.

\subsubsection{Knowledge types}
Here, we further analyze the impact of relations types in the KGV's performance. 
We conduct ablation experiments on road sign recognition, car recognition. 

As Table \ref{tab: road sign recognition w/o} demonstrates, in each study, one type of relation is excluded, i.e. sign legends, colors, and shapes. 
Our findings indicate that shape is the most critical factor in enhancing performance under the distribution shift from German to Chinese and Russian road signs. 
This is likely because road signs with the same semantic meaning share identical shapes in Germany, China, and Russia. 
Sign legends also significantly contributes to performance, while information about colors is the least impactful.

\begin{table}[ht]
\centering
\resizebox{.35\textwidth}{!}{
        \begin{tabular}{ c | c c }
        \toprule
            Model & CTSD & RTSD \\
        \midrule
            KGV & \textbf{74.8\%} & \textbf{80.1\%}\\
            KGV w/o sign legend & 73.4\% \textcolor{light-green}{(+1.4\%)} & 75.5\% \textcolor{light-green}{(+4.6\%)}\\
            KGV w/o colors & 73.7\% \textcolor{light-green}{(+1.1\%)} & 77.2\% \textcolor{light-green}{(+2.9\%)} \\
            KGV w/o shapes & 72.8\% \textcolor{light-green}{(+2.0\%)}& 74.7\% \textcolor{light-green}{(+5.4\%)} \\
        \bottomrule
        \end{tabular}
}
\caption{Classification Accuracy in the Road Sign Domain. The models are trained with 43 classes data of GTSRB and testing with shared 25 classes data of CTSD and shared 36 classes data of RTSD.}
\label{tab: road sign recognition w/o}
\end{table}

On the DVM-CAR dataset, we performed experiments  in a setting with 5\% training data. 
We observe that even with this amount of data, KGV  exhibits a large performance margin compared to the baselines. 
As indicated in Table \ref{tab: DVM-CAR dataset w/o}, the information about the body type is the most important factor in improving performance. Without this information, there is a significant drop of 4.0\%. Additionally, prior knowledge of colors is also crucial, as the performance drops by 2.4\% without color information.

We also found that non-morphometric features, such as gear type and fuel type, can also influence performance. This is likely because specific car models are often available only with certain types of gear and fuel. For instance, Tesla primarily produces electric cars, and most of them are automatic. The prior knowledge of the image viewpoint does not play a significant role.

\begin{table}[h]
\centering
\resizebox{.48\textwidth}{!}{
        \begin{tabular}{ c | c | c | c }
        \toprule
            Model & Accuracy & Model & Accuracy \\
        \midrule
            KGV & \textbf{77.1\%} & ResNet50 & 64.0\% \textcolor{light-green}{(+13.1\%)}\\
        \midrule
            KGV w/o body type & 73.1\% \textcolor{light-green}{(+4.0\%)} & KGV w/o color & 74.7\% \textcolor{light-green}{(+2.4\%)}\\
            KGV w/o viewpoint & 76.2\% \textcolor{light-green}{(+0.9\%)} & KGV w/o non-morpho. & 75.2\% \textcolor{light-green}{(+1.9\%)}\\
        \bottomrule
        \end{tabular}
}
\caption{Accuracy in DVM-CAR Model Classification with 5\% Training Data.}
\label{tab: DVM-CAR dataset w/o}
\end{table}


%% file: sections/05_conclusion.tex
\section{Conclusion and Future Work}
\label{sec:conclusion}
In this paper, we introduce KGV, a knowledge-guided visual representation learning method that leverages multi-modal prior knowledge to improve generalization in deep learning. 
The multi-modal prior knowledge across data distributions in KGV helps to mitigate overfitting the training dataset, thereby enhancing the robustness of the model against data distribution shifts. 
We observe that modeling node embeddings of the KG as Gaussian embeddings enhances the performance of KGV to model hierachical relations.
Furthermore, our method demonstrates a strong tendency to be data efficient, i.e consistently achieves good performance in the low data regimes. 
Finally, we show that our KGV method can be well adapted to the SOTA visual foundation models such as DINOv2 and CLIP. 



A main objective of our future work is to scale our concept to larger datasets and KGs.
Additionally, we plan to explore further methods for aligning KGEs and image embeddings, and inducing information of the KG at inference time. Lastly, we will look whether incorporating additional text information will improve the performance.


%% file: sections/06_appendix.tex
\clearpage
The following sections provide detailed information about the datasets and the training process for each scenario in our experiments. Additionally, we present further analysis through ablation experiments.

\section{Dataset Settings}

Detailed descriptions of the used datasets are provided in this section.

\subsection{Road Sign:}
\input{tikz/hierarchical_relations2}
We leverage three commonly used road sign recognition datasets to evaluate our model. They are the German Traffic Sign Recognition Benchmark (GTSRB) from \citet{6033395}, Chinese Traffic Sign Dataset (CTSD) from \citet{yang2015towards}, and Russian Traffic Sign Images Dataset (RTSD) from \citet{shakhuro2016russian}. The detailed information on these three datasets and the synthetic images of road sign elements are listed in Table~\ref{tab:road sign recognition datasets} and Table~\ref{tab:feature images}. The hierarchical relations are shown in Figure~\ref{fig:hierachical relations}.
\begin{table}[ht]
    \centering
    \begin{tabular}{c c c}
        \toprule
        Dataset & Total Images & Road Sign Classes \\
        \midrule
        GTSRB & 51970 & 43 \\

        CTSD & 6164 & 58 \\
        
        RTSD & 32983 & 64 \\
        \bottomrule
    \end{tabular}
    \caption{GTSRB, CTSD, and RTSD represent three road sign recognition datasets of China, Germany, and Russia. The total images and class numbers are listed in the table. }
    \label{tab:road sign recognition datasets}
\end{table}
\begin{table}[!ht]
    \centering
    \begin{tabular}{m{1.3cm} | m{1.5cm} m{1cm} | m{1.5cm} m{1cm}}
        \toprule
        Feature Types & Sub-class & Image & Sub-class & Image\\
        \midrule
        & Black & \includegraphics[width=0.6cm,height=0.6cm]{images/Color_Black.png} & Blue & \includegraphics[width=0.6cm,height=0.6cm]{images/Color_Blue.png} \\ 
        Color & Brown & \includegraphics[width=0.6cm,height=0.6cm]{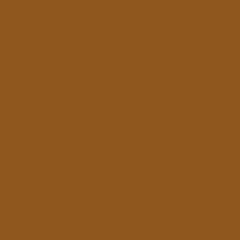} & Green &  \includegraphics[width=0.6cm,height=0.6cm]{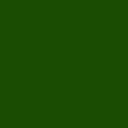} \\ 
        & Red & \includegraphics[width=0.6cm,height=0.6cm]{images/Color_Red.png} & White &  \includegraphics[width=0.6cm,height=0.6cm]{images/Color_White.png} \\ 
        & Yellow & \includegraphics[width=0.6cm,height=0.6cm]{images/Color_Yellow.png} &  &  \\ 
        \midrule 
        & Circle & \includegraphics[width=0.6cm,height=0.6cm]{images/Shape_Circle.png} & Diamond & \includegraphics[width=0.6cm,height=0.6cm]{images/Shape_Diamond.png} \\
        Shape & Octagon & \includegraphics[width=0.6cm,height=0.6cm]{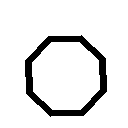} & Triangle-Down & \includegraphics[width=0.6cm,height=0.6cm]{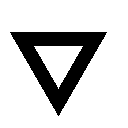} \\
        & Triangle-Up &\includegraphics[width=0.6cm,height=0.6cm]{images/Shape_TriangleUp.png} &  &  \\
        \midrule
        & Animal & \includegraphics[width=0.6cm,height=0.6cm]{images/Icon_Animal.png} & Bicycle & \includegraphics[width=0.6cm,height=0.6cm]{images/Icon_Bicycle.png} \\
        & Exclamation & \includegraphics[width=0.6cm,height=0.6cm]{images/Icon_Exclamation.png} & Person & \includegraphics[width=0.6cm,height=0.6cm]{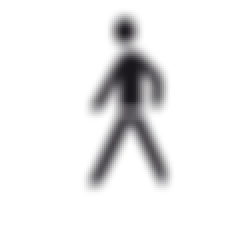} \\
        Sign Legend& Sand &\includegraphics[width=0.6cm,height=0.6cm]{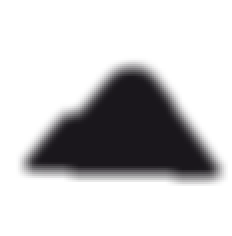} & SnowFlake  &  \includegraphics[width=0.6cm,height=0.6cm]{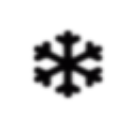}\\
        & Truck  &\includegraphics[width=0.6cm,height=0.6cm]{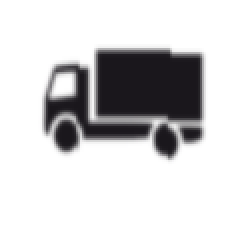} & TwoPerson  &  \includegraphics[width=0.6cm,height=0.6cm]{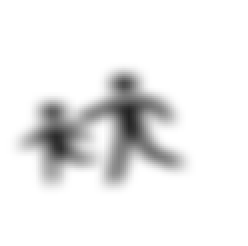}\\
        & Vehicle  &\includegraphics[width=0.6cm,height=0.6cm]{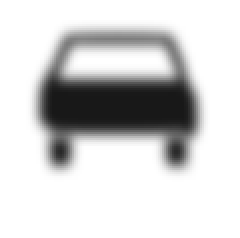} &  & \\

        \bottomrule
    \end{tabular}
    \caption{Synthetic Images in the Road Sign Recognition Domain. Seven colors, five shapes, and nine sign legends are included. For each type, we generated synthetic images as illustrated in the table.}
    \label{tab:feature images}
\end{table}

\subsection{DVM-CAR:}
\begin{table}[!ht]
    \centering
    \begin{tabular}{m{1.3cm} | m{1.5cm} m{1cm} | m{1.5cm} m{1cm}}
        \toprule
        Feature Types & Sub-class & Image & Sub-class & Image\\
        \midrule
        & Beige & \includegraphics[width=0.6cm,height=0.6cm]{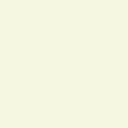} & Black & \includegraphics[width=0.6cm,height=0.6cm]{images/Color_Black.png} \\ 
        & Blue & \includegraphics[width=0.6cm,height=0.6cm]{images/Color_Blue.png} & Bronze &  \includegraphics[width=0.6cm,height=0.6cm]{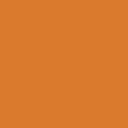} \\ 
        & Brown & \includegraphics[width=0.6cm,height=0.6cm]{images/Color_Brown.png} & Burgundy &  \includegraphics[width=0.6cm,height=0.6cm]{images/Color_Black.png} \\ 
        & Gold & \includegraphics[width=0.6cm,height=0.6cm]{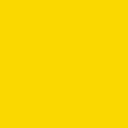} & Green &  \includegraphics[width=0.6cm,height=0.6cm]{images/Color_Green.png} \\ 
        & Grey & \includegraphics[width=0.6cm,height=0.6cm]{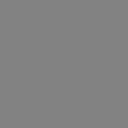} & Indigo &  \includegraphics[width=0.6cm,height=0.6cm]{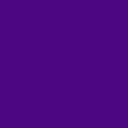} \\ 
        Color & Magenta & \includegraphics[width=0.6cm,height=0.6cm]{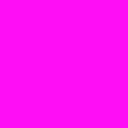} & Maroon &  \includegraphics[width=0.6cm,height=0.6cm]{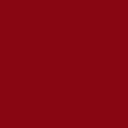} \\ 
        & Navy & \includegraphics[width=0.6cm,height=0.6cm]{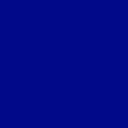} & Orange &  \includegraphics[width=0.6cm,height=0.6cm]{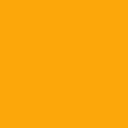} \\ 
        & Pink & \includegraphics[width=0.6cm,height=0.6cm]{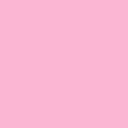} & Purple &  \includegraphics[width=0.6cm,height=0.6cm]{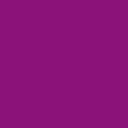} \\ 
        & Red & \includegraphics[width=0.6cm,height=0.6cm]{images/Color_Red.png} & Silver &  \includegraphics[width=0.6cm,height=0.6cm]{images/Color_Silver.png} \\
        & Turquoise & \includegraphics[width=0.6cm,height=0.6cm]{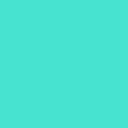} & White &  \includegraphics[width=0.6cm,height=0.6cm]{images/Color_White2.png} \\

        & Yellow & \includegraphics[width=0.6cm,height=0.6cm]{images/Color_Yellow.png} &  & \\
        \midrule
        & Convertible & \includegraphics[width=0.8cm,height=0.6cm]{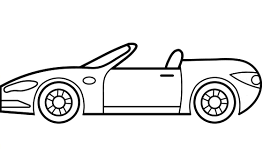} & Coupe & \includegraphics[width=0.8cm,height=0.6cm]{images/Bodytype_Coupe.png} \\
        & Estate & \includegraphics[width=0.8cm,height=0.6cm]{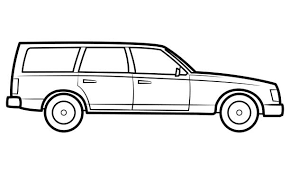} & Hatchback & \includegraphics[width=0.8cm,height=0.6cm]{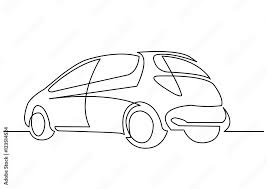} \\
        Bodytype & Limousine &\includegraphics[width=0.8cm,height=0.6cm]{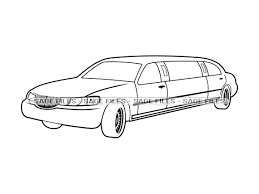} & Minibus & \includegraphics[width=0.8cm,height=0.6cm]{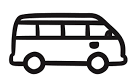}  \\
        & MPV & \includegraphics[width=0.6cm,height=0.6cm]{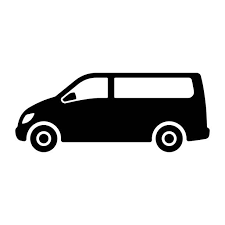} & Pickup & \includegraphics[width=0.8cm,height=0.6cm]{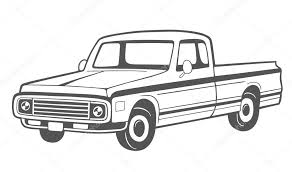} \\
        & Saloon & \includegraphics[width=0.8cm,height=0.6cm]{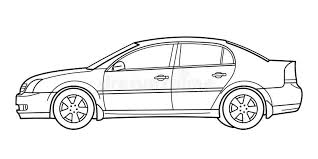} & SUV & \includegraphics[width=0.8cm,height=0.6cm]{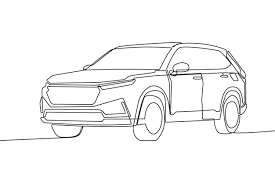} \\
        & Van &\includegraphics[width=0.8cm,height=0.6cm]{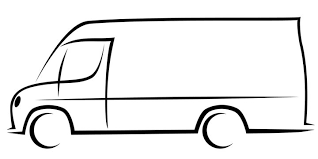} &  &  \\
        \bottomrule
    \end{tabular}
    \caption{Synthetic Images in the Car Recognition Domain. Twenty-one colors and eleven body type sketches are included. We generate synthetic images for each color and collect body type sketches on the internet.}
    \label{tab:feature images DVM}
\end{table}
Deep Visual Marketing Car (DVM-CAR) dataset~\cite{DBLP:journals/corr/abs-2109-00881} is created from 335,562 used car advertisements. It contains 1,451,784 images of cars from various angles, as well as their sales and technical data. We use this dataset to predict the car model from the given images. All images are resized to the resolution of 128 $\times$ 128. We use the same training, validation, and testing setting as MMCL~\cite{Hager_2023_CVPR}. Their dataset consists of 286 classes with a total of 418,100 training samples, 104,308 validation samples, and 525,557 testing samples. Samples of synthetic images used for training can be seen in Table \ref{tab:feature images DVM}. We generate 20 pure color images with small variations for each color category. For each body type category, we collect 20 sketch images. Note that we extract the image instance-level triplets from the tabular annotations provided in the DVM-CAR dataset. Figure \ref{fig:association relations dvm} shows an abstract of relations between one sample image instance and the other nodes in the KG. 
\input{tikz/association_relations_DVM}

\subsection{ImageNet:}
\input{tikz/hierarchical_relation2_imagenet}
Mini-ImageNet \cite{vinyals2016matching} is a subset of the ImageNet dataset, consisting of 60K images of size $84 \times 84$ with 100 classes, each having 600 examples. ImageNet-Skecth~\cite{wang2019learning} comprises 50000 images, 50 images for each of the 1000 ImageNet classes. Here, we collect the 100 classes in mini-ImageNet from 1000 classes of ImageNet-Sketch for training. 
We employ three distinct datasets for our evaluation: ImageNetV2~\cite{recht2019imagenet}, ImageNet-R~\cite{hendrycks2021many}, and ImageNet-A~\cite{hendrycks2021natural}. ImageNetV2 comprises 10 fresh test images for each class, sharing the same set of labels as mini-ImageNet. ImageNet-R includes various artistic representations such as art, cartoons, and deviant styles, covering 200 classes from ImageNet. Lastly, ImageNet-A features unaltered, real-world examples that occur naturally. For our purposes, we assess only those classes that overlap with mini-ImageNet. An abstract of hierarchical relations extracted from the WordNet in the ImageNet Domain is illustrated in Figure \ref{fig:hierarchical relations imagenet}.

\section{Training Details}
\label{sec: training details}
Our baseline ResNet50 has a simple structure of a ResNet50 backbone as an encoder and full connect layers as a decoder. The ResNet50 backbone is untrained and randomly initialized in a Xavier manner. Our KGV model and the other baselines, GCNZ and DGP, use the same image encoder backbone. The node and relation lookup tables, which contain the embeddings of nodes and relations in the knowledge graph, are also initialized, following the initialization method proposed in transE \cite{NIPS2013_1cecc7a7}. The relation, represented as `sub class of/instance of' is an exception. It is initialized with a vector of all zeros and frozen during the training procedure. We use one NVIDIA A100 GPU with 80GB memory to train the models. Each algorithm is run with three random seeds to get the average performance. 

The hyperparameters used in the three domains are illustrated as follows: 
\begin{itemize}
    \item \textbf{Road Sign Recognition Domain}: The weight balance $\beta$ defined is 0.1. We train the baseline and our model with the Adam optimizer and a constant learning rate of 0.0002. The total training epochs are 120. The mini-batch size is 32. We resize all the input images to $32 \times 32$.
    \item \textbf{Car Recognition Domain}: We resize all input image to 128 $\times$ 128. The batch size we select is 512. Adam optimizer with a learning rate of 0.001 and cosines annealing learning rate schedule is utilized for optimization. The training epochs for 5\%, 10\%, 50\%, and 100\% of data are 600, 200, 150, and 100, respectively. The balance $\beta$ we use is 0.5. 
    \item \textbf{ImageNet Domain}: Both Baseline and our model KGV are trained with Adam optimizer and a learning rate of 0.001. The batch size we select is 1024 and the training epoch is 150. We resize all the input images to $84 \times 84$. The balance $\beta$ we use is 0.5. 
\end{itemize}

\section{Additional Analysis}
In this section, we further analyze which knowledge plays the most important role in the performance.

\subsection{ImageNet:}
\begin{table}[ht]
\centering
\resizebox{.42\textwidth}{!}{
    \begin{tabular}{ c | c  c c}
    \toprule
        Model & KGV & \makecell{KGV \\ w/o color} & \makecell{KGV \\ w/o sketch} \\
    \midrule
        Mini-ImageNet & \textbf{66.8\%} & \makecell{63.6\%\\ \textcolor{light-green}{+3.2\%} } &  \makecell{64.4\%\\ \textcolor{light-green}{+2.4\%} }\\
        ImageNetV2 & \textbf{52.7\%} &  \makecell{50.0\%\\ \textcolor{light-green}{+2.7\%} } &  \makecell{51.5\%\\ \textcolor{light-green}{+1.2\%} }\\
        ImageNet-R & \textbf{41.4\%} &  \makecell{39.5\%\\ \textcolor{light-green}{+1.9\%} } &  \makecell{29.7\%\\ \textcolor{light-green}{+11.7\%} }\\
        ImageNet-A & \textbf{6.8\%} &  \makecell{5.1\%\\ \textcolor{light-green}{+1.7\%} }  & \makecell{6.2\%\\ \textcolor{light-green}{+0.6\%} }\\
    \bottomrule
    \end{tabular}
}
\caption{Classification Accuracy in the ImageNet Domain.}
\label{tab: ImageNet w/o}
\end{table}
As demonstrated in Table \ref{tab: ImageNet w/o}, we performed ablation studies on our KGV method by excluding either prior knowledge of color or sketch. Our results indicate that knowledge of color significantly impacts performance on the Mini-ImageNet, ImageNetV2, and ImageNet-A datasets. The knowledge of sketch proves to be more critical for classifying the ImageNet-R dataset, likely due to the similarity between the ImageNet-Sketch and ImageNet-R datasets.

%% file: tikz/hierarchical_relations2.tex
\tikzstyle{arrow} = [very thick,->,>=stealth]
\begin{figure}[ht]
    \centering

    \begin{adjustbox}{width=0.44\textwidth}
    \begin{tikzpicture}[node distance=2cm]
        \node[circle, draw=light-orange, very thick, minimum height=0.4cm, label={[label distance=0cm, font=\bfseries]0: \small Road Sign}](road-sign){};
        \node[circle, draw=light-orange, very thick, right = 2cm of road-sign, yshift=2.0cm, minimum height=0.4cm, label={[label distance=0cm, font=\bfseries]0: \small Informative}](informative){};
        \node (ellipsis-informative)[right = 0.4cm of informative,yshift=0.6cm]{\Large ...};
        \node[circle, draw=light-orange, very thick, right = 2cm of road-sign, yshift=1.0cm, minimum height=0.4cm, label={[label distance=0cm, font=\bfseries]0: \small Mandatory}](mandatory){};
        \node (ellipsis-mandatory)[right = 0.4cm of mandatory,yshift=-0.6cm]{\Large ...};
        \node[circle, draw=light-orange, very thick, right = 2cm of road-sign, yshift=-3.5cm, minimum height=0.4cm, label={[label distance=0cm, font=\bfseries]0: \small Prohibitory}](prohibitory){}; 
        \node (ellipsis-prohibitory)[right = 2cm of prohibitory]{\Large ...};  
        \node[circle, draw=light-orange, very thick, right = 2cm of prohibitory, yshift=0.6cm, minimum height=0.4cm, label={[label distance=0cm, font=\bfseries]0: \small No Pedestrian}](prohibition){};
        \node (ellipsis-prohibition)[right = 0.4cm of prohibition,yshift=-0.6cm]{\Large ...};     
        \node[circle, draw=light-orange, very thick, right = 2cm of prohibitory, yshift=-0.6cm, minimum height=0.4cm, label={[label distance=0cm, font=\bfseries]0: \small Speed Limit 30}](speed-limit){};
        \node (ellipsis-speed-limit)[right = 0.4cm of speed-limit, yshift=-0.6cm]{\Large ...};     
        \node[circle, draw=light-orange, very thick, right = 2.0cm of road-sign, yshift=-1.5cm, minimum height=0.4cm, label={[label distance=0cm, font=\bfseries]0: \small Warning}](warning){};
        \node (ellipsis-warning)[right = 2cm of warning]{\Large ...};  
        \node[circle, draw=light-orange, very thick, right = 2.0cm of warning, yshift=0.6cm, minimum height=0.4cm, label={[label distance=0.3cm, font=\bfseries, text width=1.5cm]0: \small Warning Pedestrian}](warning-pedestrian){};
        \node[circle, draw=light-orange, very thick, right = 2.0cm of warning-pedestrian, yshift=0.6cm, minimum height=0.4cm, label={[label distance=0cm, font=\bfseries, text width=1.5cm]0: \small DE}](warning-pedestrian-de){};
        \node[circle, draw=light-orange, very thick, right = 2.0cm of warning-pedestrian, yshift=-0.6cm, minimum height=0.4cm, label={[label distance=0cm, font=\bfseries, text width=1.5cm]0: \small CN}](warning-pedestrian-cn){};
        \node[circle, draw=light-orange, very thick, right = 2.0cm of warning-pedestrian, yshift=1.2cm, minimum height=0.4cm, label={[label distance=0cm, font=\bfseries, text width=1.5cm]0: \small RS}](warning-pedestrian-rs){};
        
        \node[circle, draw=light-orange, very thick, right = 2.0cm of warning, yshift=-0.6cm, minimum height=0.4cm, label={[label distance=0cm, font=\bfseries, text width=1.5cm]0: \small Warning Animal}](warning-animal){};
        \node[circle, draw=light-orange, very thick, minimum height=0.4cm, below = 6cm of road-sign, xshift=0cm, label={[label distance=0cm, font=\bfseries, text width=2cm]0: \small Road Sign Feature}](road-sign-feature){};
        \node[circle, draw=light-orange, very thick, minimum height=0.4cm, right = 2cm of road-sign-feature, yshift=0.8cm, label={[label distance=0cm, font=\bfseries]0: \small Shape}](shape){};
        \node (ellipsis-shape)[right = 2cm of shape]{\Large ...};
        \node[circle, draw=light-orange, very thick, minimum height=0.4cm, right = 2cm of shape, yshift=0.4cm, label={[label distance=0cm, font=\bfseries]0: \small Circle}](circle){};
        \node[circle, draw=light-orange, very thick, minimum height=0.4cm, right = 2cm of shape, yshift=-0.4cm, label={[label distance=0cm, font=\bfseries]0: \small Triangle}](triangle){};

        \node[circle, draw=light-orange, very thick, minimum height=0.4cm, right = 2cm of road-sign-feature, yshift=-0.8cm, label={[label distance=0cm, font=\bfseries]0: \small Sign Legend}](legend){};
        \node (ellipsis-legend)[right = 2cm of legend]{\Large ...};
        \node[circle, draw=light-orange, very thick, minimum height=0.4cm, right = 2cm of legend, yshift=0.4cm, label={[label distance=0cm, font=\bfseries]0: \small Exclamation}](exclamation){};
        \node[circle, draw=light-orange, very thick, minimum height=0.4cm, right = 2cm of legend, yshift=-0.4cm, label={[label distance=0cm, font=\bfseries]0: \small Truck}](truck){};
        \node[circle, draw=light-orange, very thick, minimum height=0.4cm, right = 2cm of road-sign-feature, yshift=-2.4cm, label={[label distance=0cm, font=\bfseries]0: \small Color}](color){};
        \node (ellipsis-color)[right = 2cm of color]{\Large ...};
        \node[circle, draw=light-orange, very thick, minimum height=0.4cm, right = 2cm of color, yshift=-0.4cm, label={[label distance=0cm, font=\bfseries]0: \small Red}](red){};
        \node[circle, draw=light-orange, very thick, minimum height=0.4cm, right = 2cm of color, yshift=0.4cm, label={[label distance=0cm, font=\bfseries]0: \small White}](white){};
        \draw[draw=light-purple, arrow, dashed] (informative) to [out=180,in=90] (road-sign);
        \draw[draw=light-purple, arrow, dashed] (mandatory) to [out=180,in=45] (road-sign);
        \draw[draw=light-purple, arrow, dashed] (prohibitory) to [out=180,in=-90] (road-sign);
        \draw[draw=light-purple, arrow, dashed] (warning) to [out=180,in=-45] (road-sign);
        \draw[draw=light-purple, arrow, dashed] (prohibition) to [out=180,in=40] (prohibitory);
        \draw[draw=light-purple, arrow, dashed] (speed-limit) to [out=180,in=-40] (prohibitory);
        \draw[draw=light-purple, arrow, dashed] (shape) to [out=180,in=60] (road-sign-feature);
        \draw[draw=light-purple, arrow, dashed] (legend) to [out=180,in=-60] (road-sign-feature);
        \draw[draw=light-purple, arrow, dashed] (color) to [out=180,in=-80] (road-sign-feature);
        \draw[draw=light-purple, arrow, dashed] (warning-pedestrian) to [out=180,in=45] (warning);
        \draw[draw=light-purple, arrow, dashed] (warning-animal) to [out=180,in=-45] (warning);
        \draw[draw=light-purple, arrow, dashed] (warning-pedestrian-de) to [out=180,in=45] (warning-pedestrian);
        \draw[draw=light-purple, arrow, dashed] (warning-pedestrian-cn) to [out=180,in=-45] (warning-pedestrian);
        \draw[draw=light-purple, arrow, dashed] (warning-pedestrian-rs) to [out=180,in=60] (warning-pedestrian);
        \draw[draw=light-purple, arrow, dashed] (ellipsis-informative) to [out=180,in=60] (informative);
        \draw[draw=light-purple, arrow, dashed] (ellipsis-prohibition) to [out=180,in=-60] (prohibition);
        \draw[draw=light-purple, arrow, dashed] (ellipsis-speed-limit) to [out=180,in=-60] (speed-limit);
        \draw[draw=light-purple, arrow, dashed] (ellipsis-mandatory) to [out=180,in=-60] (mandatory);
        \draw[draw=light-purple, arrow, dashed] (circle) to [out=180,in=40] (shape);
        \draw[draw=light-purple, arrow, dashed] (triangle) to [out=180,in=-40] (shape);
        \draw[draw=light-purple, arrow, dashed] (exclamation) to [out=180,in=40] (legend);
        \draw[draw=light-purple, arrow, dashed] (truck) to [out=180,in=-40] (legend);
        \draw[draw=light-purple, arrow, dashed] (white) to [out=180,in=40] (color);
        \draw[draw=light-purple, arrow, dashed] (red) to [out=180,in=-40] (color);
    \end{tikzpicture}
    \end{adjustbox}
    \caption{An abstract of hierarchical relations existing in the road sign recognition domain.}
    \label{fig:hierachical relations}
\end{figure}

%% file: tikz/association_relations_DVM.tex
\tikzstyle{arrow} = [thick,->,>=stealth]
\tikzstyle{myshadow2} = [drop shadow={
            shadow scale=0.98,
            shadow xshift=0.5ex,
            shadow yshift=-0.5ex
        }]
\begin{figure}[ht]
    \centering
    \begin{tikzpicture}[node distance=2cm]
        \node[inner sep=0pt, myshadow2, draw=black] (porsche){\includegraphics[width=1.5cm]{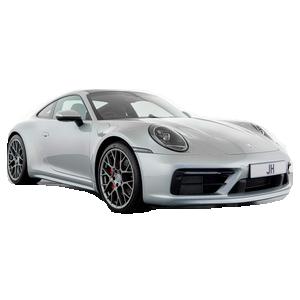}};

        \node (color) [circle, minimum size=0.4cm, draw=light-orange, thick, right of = porsche, xshift=0.5cm, yshift=2cm, label={[label distance=0cm, font=\bfseries]90: \scriptsize Color}]{};
        \node (silver) [circle, minimum size=0.4cm, draw=light-orange, thick, below = 0.3cm of color, xshift=-0.5cm, label={[label distance=0cm, font=\bfseries]100: \scriptsize Silver}]{};
        \node (red) [circle, minimum size=0.4cm, draw=light-orange, thick, below = 0.3cm of color, xshift=0.5cm, label={[label distance=0cm, font=\bfseries]80: \scriptsize Red}]{};
        \node (ellipsis-color)[right = 0.0cm of silver, yshift=-0.1cm]{\Large ...};
        \draw [arrow, dashed, draw=light-purple] (silver) -- (color);        \draw [arrow, dashed, draw=light-purple] (red) -- (color);

        \node (body_type) [circle, minimum size=0.4cm, draw=light-orange, thick, right of = porsche, xshift=1.0cm, yshift=0cm, label={[label distance=0cm, font=\bfseries]90: \scriptsize Body Type}]{};
        \node (coupe) [circle, minimum size=0.4cm, draw=light-orange, thick, below = 0.3cm of body_type, xshift=-0.5cm, label={[label distance=0cm, font=\bfseries, xshift=-0.2cm]90: \scriptsize Coupe}]{};
        \node (suv) [circle, minimum size=0.4cm, draw=light-orange, thick, below = 0.3cm of body_type, xshift=0.5cm, label={[label distance=0cm, font=\bfseries]80: \scriptsize SUV}]{};
        \node (ellipsis-body-type)[right = 0.0cm of coupe, yshift=-0.1cm]{\Large ...};
        \draw [arrow, dashed, draw=light-purple] (coupe) -- (body_type);        \draw [arrow, dashed, draw=light-purple] (suv) -- (body_type);

        \node (gear_type) [circle, minimum size=0.4cm, draw=light-orange, thick, left of = porsche, xshift=0cm, yshift=2.5cm, label={[label distance=0cm, font=\bfseries]90: \scriptsize Gear Type}]{};
        \node (auto) [circle, minimum size=0.4cm, draw=light-orange, thick, below = 0.3cm of gear_type, xshift=-0.5cm, label={[label distance=0cm, font=\bfseries]95: \scriptsize Auto}]{};
        \node (manuel) [circle, minimum size=0.4cm, draw=light-orange, thick, below = 0.3cm of gear_type, xshift=0.5cm, label={[label distance=0cm, font=\bfseries]80: \scriptsize Manuel}]{};
        \node (ellipsis-gear-type)[right = 0.0cm of auto, yshift=-0.1cm]{\Large ...};
        \draw [arrow, dashed, draw=light-purple] (auto) -- (gear_type);        \draw [arrow, dashed, draw=light-purple] (manuel) -- (gear_type);

        \node (fuel_type) [circle, minimum size=0.4cm, draw=light-orange, thick, left of = porsche, xshift=-1cm, yshift=0.5cm, label={[label distance=0cm, font=\bfseries]90: \scriptsize Fuel Type}]{};
        \node (diesel) [circle, minimum size=0.4cm, draw=light-orange, thick, below = 0.3cm of fuel_type, xshift=-0.5cm, label={[label distance=0cm, font=\bfseries]-95: \scriptsize Diesel}]{};
        \node (petrol) [circle, minimum size=0.4cm, draw=light-orange, thick, below = 0.3cm of fuel_type, xshift=0.5cm, label={[label distance=0cm, font=\bfseries]-80: \scriptsize Petrol}]{};
        \node (ellipsis-fuel-type)[right = 0.0cm of auto, yshift=-0.1cm]{\Large ...};
        \draw [arrow, dashed, draw=light-purple] (diesel) -- (fuel_type);      \draw [arrow, dashed, draw=light-purple] (petrol) -- (fuel_type);

        \node (number) [circle, minimum size=0.4cm, draw=light-orange, thick, right of = porsche, xshift=0cm, yshift=-3cm, label={[label distance=0cm, font=\bfseries]-90: \scriptsize Number}]{};
        \node (number2) [circle, minimum size=0.4cm, draw=light-orange, thick, above = 0.3cm of number, xshift=-0.5cm, label={[label distance=0cm, font=\bfseries]85: \scriptsize 2}]{};
        \node (number5) [circle, minimum size=0.4cm, draw=light-orange, thick, above = 0.3cm of number, xshift=0.5cm, label={[label distance=0cm, font=\bfseries]80: \scriptsize 5}]{};
        \node (ellipsis-number)[right = 0.0cm of number2, yshift=-0.1cm]{\Large ...};
        \draw [arrow, dashed, draw=light-purple] (number2) -- (number);      \draw [arrow, dashed, draw=light-purple] (number5) -- (number);

        \node (viewpoint) [circle, minimum size=0.4cm, draw=light-orange, thick, left of = porsche, xshift=0cm, yshift=-3cm, label={[label distance=0cm, font=\bfseries]-90: \scriptsize Viewpoint}]{};
        \node (degree0) [circle, minimum size=0.4cm, draw=light-orange, thick, above = 0.3cm of viewpoint, xshift=-0.75cm, label={[label distance=0cm, font=\bfseries]90: \scriptsize 0$^{\circ}$}]{};
        \node (degree45) [circle, minimum size=0.4cm, draw=light-orange, thick, above = 0.3cm of viewpoint, xshift=-0.125cm, label={[label distance=0cm, font=\bfseries, xshift=-0.1cm]90: \scriptsize 45$^{\circ}$}]{};
        \node (degree315) [circle, minimum size=0.4cm, draw=light-orange, thick, above = 0.3cm of viewpoint, xshift=0.875cm, label={[label distance=0cm, font=\bfseries]90: \scriptsize 315$^{\circ}$}]{};
        \node (ellipsis-viewpoint)[right = 0.0cm of degree45, yshift=-0.1cm]{\Large ...};
        \draw [arrow, dashed, draw=light-purple] (degree0) -- (viewpoint);     \draw [arrow, dashed, draw=light-purple] (degree45) -- (viewpoint);
        \draw [arrow, dashed, draw=light-purple] (degree315) -- (viewpoint);

        \draw [arrow, draw=light-purple] (porsche) to node [above,midway, color=light-purple, rotate=28] {\scriptsize hasColor} (silver);

        \draw [arrow, draw=light-purple] (porsche) to node [below,midway, color=light-purple, rotate=-18] {\scriptsize hasBodyType} (coupe);
        \draw [arrow, draw=light-purple] (porsche) to node [below,midway, color=light-purple, rotate=-58] {\scriptsize hasDoorNo} (number2);

        \draw [arrow, draw=light-purple] (porsche) to node [above,midway, color=light-purple, rotate=48] {\scriptsize hasViewpoint} (degree45);
        \draw [arrow, draw=light-purple] (porsche) to node [above,midway, color=light-purple, rotate=5] {\scriptsize hasFuelType} (petrol);
        \draw [arrow, draw=light-purple] (porsche) to node [below,midway, color=light-purple, rotate=-35] {\scriptsize hasGearType} (auto);

       \node (porsche911) [circle, minimum size=0.4cm, draw=light-orange, thick, above of = porsche, xshift=1cm, yshift=0.5cm, label={[label distance=0cm, font=\bfseries]90: \scriptsize Porsche 911}]{};

        \draw [arrow, dashed, draw=light-purple] (porsche) to node [above,midway, color=light-purple, rotate=65] {\scriptsize instanceOf} (porsche911);

    \end{tikzpicture}
    \caption{The relations between one image instance of Porsche 911 with other nodes in the car recognition KG.}
    \label{fig:association relations dvm}
    \vspace{1.5em}
\end{figure}

%% file: tikz/hierarchical_relation2_imagenet.tex
\tikzstyle{arrow} = [very thick,->,>=stealth]
\begin{figure}[h]
    \centering

    \begin{adjustbox}{width=0.45\textwidth}
    \begin{tikzpicture}[node distance=2cm]
        \node[circle, draw=light-orange, very thick, minimum height=0.4cm, label={[label distance=0cm, font=\bfseries]0: \small Thing}](thing){};
        \node (ellipsis-thing)[right = 0.4cm of thing,yshift=-0.6cm]{\Large ...};  
        \node[circle, draw=light-orange, very thick, right = 1cm of road-sign, yshift=6.0cm, minimum height=0.4cm, label={[label distance=0cm, font=\bfseries]0: \small Food}](food){};
        \node (ellipsis-food)[right = 0.2cm of food, yshift=0.4cm]{\Large ...};  
        
        \node[circle, draw=light-orange, very thick, right = 1cm of food, yshift=1.2cm, minimum height=0.4cm, label={[label distance=0cm, font=\bfseries]0: \small Beverage}](beverage){};
        \node (ellipsis-beverage)[right = 0.2cm of beverage, yshift=1.0cm]{\Large ...};  
        \node[circle, draw=light-orange, very thick, right = 1cm of beverage, yshift=0.6cm, minimum height=0.4cm, label={[label distance=0cm, font=\bfseries]0: \small Milk}](milk){};
        \node[circle, draw=light-orange, very thick, right = 1cm of beverage, yshift=-0.6cm, minimum height=0.4cm, label={[label distance=0cm, font=\bfseries]0: \small Beer}](beer){};

        \node[circle, draw=light-orange, very thick, right = 1cm of food, yshift=-1.2cm, minimum height=0.4cm, label={[label distance=0cm, font=\bfseries]0: \small Eat}](eat){};
        \node (ellipsis-eat)[right = 0.2cm of eat, yshift=1.0cm]{\Large ...};  

        \node[circle, draw=light-orange, very thick, right = 1cm of eat, yshift=0.6cm, minimum height=0.4cm, label={[label distance=0cm, font=\bfseries]0: \small Corn}](corn){};

        \node[circle, draw=light-orange, very thick, right = 1cm of eat, yshift=-0.6cm, minimum height=0.4cm, label={[label distance=0cm, font=\bfseries]0: \small Hotdog}](hotdog){};

        \node[circle, draw=light-orange, very thick, right = 1cm of road-sign, yshift=1.2cm, minimum height=0.4cm, label={[label distance=0cm, font=\bfseries]0: \small Place}](place){};
        \node (ellipsis-place)[right = 0.2cm of place, yshift=0.4cm]{\Large ...};  
        
        \node[circle, draw=light-orange, very thick, right = 1cm of place, yshift=1.2cm, minimum height=0.4cm, label={[label distance=0cm, font=\bfseries]0: \small Architecture}](architecture){};
        \node (ellipsis-architecture)[right = 0.2cm of architecture, yshift=1.0cm]{\Large ...};  

        \node[circle, draw=light-orange, very thick, right = 1cm of architecture, yshift=0.6cm, minimum height=0.4cm, label={[label distance=0cm, font=\bfseries]0: \small Building}](building){};

        \node[circle, draw=light-orange, very thick, right = 1cm of architecture, yshift=0.6cm, minimum height=0.4cm, label={[label distance=0cm, font=\bfseries]0: \small Building}](building){};
        \node (ellipsis-building)[right = 0.2cm of building, yshift=0.8cm]{\Large ...}; 

        \node[circle, draw=light-orange, very thick, right = 1cm of building, yshift=0.6cm, minimum height=0.4cm, label={[label distance=0cm, font=\bfseries]0: \small Bookshop}](bookshop){};

        \node[circle, draw=light-orange, very thick, right = 1cm of building, yshift=-0.6cm, minimum height=0.4cm, label={[label distance=0cm, font=\bfseries]0: \small Dome}](dome){};

        \node[circle, draw=light-orange, very thick, right = 1cm of architecture, yshift=-0.6cm, minimum height=0.4cm, label={[label distance=0cm, font=\bfseries]0: \small Space}](space){};
        \node (ellipsis-space)[right = 0.2cm of space, yshift=-0.4cm]{\Large ...}; 
        
        \node[circle, draw=light-orange, very thick, right = 1cm of place, yshift=-1.2cm, minimum height=0.4cm, label={[label distance=0cm, font=\bfseries]0: \small NaturalSpace}](naturalspace){};
        \node (ellipsis-naturalspace)[right = 0.2cm of naturalspace, yshift=1.0cm]{\Large ...};  

        \node[circle, draw=light-orange, very thick, right = 1cm of naturalspace, yshift=0.6cm, minimum height=0.4cm, label={[label distance=0cm, font=\bfseries]0: \small Cliff}](cliff){};

        \node[circle, draw=light-orange, very thick, right = 1cm of naturalspace, yshift=-0.6cm, minimum height=0.4cm, label={[label distance=0cm, font=\bfseries]0: \small CoralReef}](coral_reef){};

        \node[circle, draw=light-orange, very thick, right = 1cm of thing, yshift=-3cm, minimum height=0.4cm, label={[label distance=0cm, font=\bfseries]0: \small Species}](species){};
        \node (ellipsis-species)[right = 0.2cm of species, yshift=-1.0cm]{\Large ...};  
        
        \node[circle, draw=light-orange, very thick, right = 1cm of species, yshift=1.6cm, minimum height=0.4cm, label={[label distance=0cm, font=\bfseries]0: \small Archaea}](archaea){};
        \node (ellipsis-archaea)[right = 0.2cm of archaea, yshift=0.4cm]{\Large ...}; 
        
        \node[circle, draw=light-orange, very thick, right = 1cm of species, yshift=0.8cm, minimum height=0.4cm, label={[label distance=0cm, font=\bfseries]0: \small Bacteria}](bacteria){};
        \node (ellipsis-bacteria)[right = 0.2cm of bacteria, yshift=0.4cm]{\Large ...}; 
        
        \node[circle, draw=light-orange, very thick, right = 1cm of species, yshift=-0.6cm, minimum height=0.4cm, label={[label distance=0cm, font=\bfseries]0: \small Eukaryote}](eukaryote){};
        \node (ellipsis-eukaryote)[right = 0.2cm of eukaryote, yshift=-1.5cm]{\Large ...}; 
        
        \node[circle, draw=light-orange, very thick, right = 1cm of eukaryote, yshift=0.6cm, minimum height=0.4cm, label={[label distance=0cm, font=\bfseries]0: \small Animal}](animal){};
        \node (ellipsis-animal)[right = 0.2cm of animal, yshift=0.8cm]{\Large ...}; 

        \node[circle, draw=light-orange, very thick, right = 1cm of animal, yshift=0.6cm, minimum height=0.4cm, label={[label distance=0cm, font=\bfseries]0: \small Bird}](bird){};
        \node (ellipsis-bird)[right = 0.2cm of bird, yshift=0.4cm]{\Large ...};  
        
        \node[circle, draw=light-orange, very thick, right = 1cm of animal, yshift=-0.6cm, minimum height=0.4cm, label={[label distance=0cm, font=\bfseries]0: \small Mamma}](mamma){};
        \node (ellipsis-mamma)[right = 0.2cm of mamma, yshift=-1.0cm]{\Large ...};  

        \node[circle, draw=light-orange, very thick, right = 1cm of mamma, yshift=0.6cm, minimum height=0.4cm, label={[label distance=0cm, font=\bfseries]0: \small Cat}](cat){};
        \node (ellipsis-cat)[right = 0.2cm of cat, yshift=0.4cm]{\Large ...};  

        \node[circle, draw=light-orange, very thick, right = 1cm of mamma, yshift=-0.6cm, minimum height=0.4cm, label={[label distance=0cm, font=\bfseries]0: \small Dog}](dog){};
        \node (ellipsis-dog)[right = 0.2cm of dog, yshift=-1cm]{\Large ...};  

        \node[circle, draw=light-orange, very thick, right = 1cm of dog, yshift=0.6cm, minimum height=0.4cm, label={[label distance=0cm, font=\bfseries]0: \small Boxer}](boxer){};

        \node[circle, draw=light-orange, very thick, right = 1cm of dog, yshift=-0.6cm, minimum height=0.4cm, label={[label distance=0cm, font=\bfseries]0: \small Saluki}](saluki){};
        
        \node[circle, draw=light-orange, very thick, right = 1cm of eukaryote, yshift=-0.6cm, minimum height=0.4cm, label={[label distance=0cm, font=\bfseries]0: \small Fungus}](fungus){};
        \node (ellipsis-fungus)[right = 0.2cm of fungus, yshift=-0.4cm]{\Large ...}; 
        
        \node[circle, draw=light-orange, very thick, right = 1cm of eukaryote, yshift=-1.4cm, minimum height=0.4cm, label={[label distance=0cm, font=\bfseries]0: \small Plant}](plant){};
        \node (ellipsis-plant)[right = 0.2cm of plant, yshift=-0.4cm]{\Large ...}; 
        \node[circle, draw=light-orange, very thick, right = 1cm of thing, yshift=-6.5cm, minimum height=0.4cm, label={[label distance=0cm, font=\bfseries]0: \small Tool}](tool){};        \node (ellipsis-tool)[right = 0.2cm of tool, yshift=-1.0cm]{\Large ...};  
        
        \node[circle, draw=light-orange, very thick, right = 1cm of tool, yshift=0.6cm, minimum height=0.4cm, label={[label distance=0cm, font=\bfseries]0: \small HairSlide}](hair_slide){};
        \node (ellipsis-hair_slide)[right = 0.2cm of hair_slide, yshift=0.4cm]{\Large ...};  

        \node[circle, draw=light-orange, very thick, right = 1cm of tool, yshift=-0.6cm, minimum height=0.4cm, label={[label distance=0cm, font=\bfseries]0: \small Device}](device){};
        \node (ellipsis-device)[right = 0.2cm of device, yshift=0.6cm]{\Large ...};  
        
        \node[circle, draw=light-orange, very thick, right = 1cm of device, yshift=-0.6cm, minimum height=0.4cm, label={[label distance=0cm, font=\bfseries]0: \small Camera}](camera){};
        \node (ellipsis-camera)[right = 0.2cm of camera, yshift=1.0cm]{\Large ...};  

        \node[circle, draw=light-orange, very thick, right = 1cm of camera, yshift=0.6cm, minimum height=0.4cm, label={[label distance=0cm, font=\bfseries]0: \small Instrument}](instrument){};

        \node[circle, draw=light-orange, very thick, right = 1cm of camera, yshift=-0.6cm, minimum height=0.4cm, label={[label distance=0cm, font=\bfseries]0: \small Organ}](organ){};

        \node[circle, draw=light-orange, very thick, right = 1cm of thing, yshift=-9cm, minimum height=0.4cm, label={[label distance=0cm, font=\bfseries]0: \small Vehicle}](vehicle){};
        \node (ellipsis-vehicle)[right = 0.2cm of vehicle, yshift=-1.0cm]{\Large ...};  
        
        \node[circle, draw=light-orange, very thick, right = 1cm of vehicle, yshift=0.6cm, minimum height=0.4cm, label={[label distance=0cm, font=\bfseries]0: \small Airplane}](airplane){};
        \node (ellipsis-airplane)[right = 0.2cm of airplane, yshift=0.4cm]{\Large ...};  
        
        \node[circle, draw=light-orange, very thick, right = 1cm of vehicle, yshift=-0.6cm, minimum height=0.4cm, label={[label distance=0cm, font=\bfseries]0: \small Ship}](ship){};
        \node (ellipsis-ship)[right = 0.2cm of ship, yshift=-1cm]{\Large ...};  
        
        \node[circle, draw=light-orange, very thick, right = 1cm of ship, yshift=0.6cm, minimum height=0.4cm, label={[label distance=0cm, font=\bfseries]0: \small AircraftCarrier}](aircraft_carrier){};

        \node[circle, draw=light-orange, very thick, right = 1cm of ship, yshift=-0.6cm, minimum height=0.4cm, label={[label distance=0cm, font=\bfseries]0: \small Yawl}](yawl){};
        \draw[draw=light-purple, arrow, dashed] (ellipsis-thing) to [out=180,in=-60] (thing);
        
        \draw[draw=light-purple, arrow, dashed] (food) to [out=230,in=90] (thing);
        \draw[draw=light-purple, arrow, dashed] (ellipsis-food) to [out=180,in=40] (food);
        \draw[draw=light-purple, arrow, dashed] (beverage) to [out=-170,in=70] (food);
        \draw[draw=light-purple, arrow, dashed] (eat) to [out=170,in=-70] (food);
        \draw[draw=light-purple, arrow, dashed] (milk) to [out=-180,in=60] (beverage);
        \draw[draw=light-purple, arrow, dashed] (beer) to [out=180,in=-60] (beverage);
        \draw[draw=light-purple, arrow, dashed] (ellipsis-beverage) to [out=190,in=70] (beverage);
        \draw[draw=light-purple, arrow, dashed] (corn) to [out=180,in=60] (eat);
        \draw[draw=light-purple, arrow, dashed] (hotdog) to [out=180,in=-60] (eat);
        \draw[draw=light-purple, arrow, dashed] (ellipsis-eat) to [out=190,in=70] (eat);
        
        \draw[draw=light-purple, arrow, dashed] (place) to [out=180,in=65] (thing);
        \draw[draw=light-purple, arrow, dashed] (ellipsis-place) to [out=180,in=40] (place);
        \draw[draw=light-purple, arrow, dashed] (architecture) to [out=-170,in=70] (place);
        \draw[draw=light-purple, arrow, dashed] (naturalspace) to [out=170,in=-70] (place);
        \draw[draw=light-purple, arrow, dashed] (building) to [out=180,in=60] (architecture);
        \draw[draw=light-purple, arrow, dashed] (space) to [out=180,in=-60] (architecture);
        \draw[draw=light-purple, arrow, dashed] (ellipsis-space) to [out=180,in=-40] (space);
        \draw[draw=light-purple, arrow, dashed] (ellipsis-architecture) to [out=190,in=70] (architecture);
        \draw[draw=light-purple, arrow, dashed] (bookshop) to [out=180,in=60] (building);
        \draw[draw=light-purple, arrow, dashed] (ellipsis-building) to [out=190,in=70] (building);
        \draw[draw=light-purple, arrow, dashed] (dome) to [out=180,in=-60] (building);
        \draw[draw=light-purple, arrow, dashed] (cliff) to [out=180,in=60] (naturalspace);
        \draw[draw=light-purple, arrow, dashed] (coral_reef) to [out=180,in=-60] (naturalspace);
        \draw[draw=light-purple, arrow, dashed] (ellipsis-naturalspace) to [out=190,in=70] (naturalspace);
        
        \draw[draw=light-purple, arrow, dashed] (species) to [out=150,in=-75] (thing);
        \draw[draw=light-purple, arrow, dashed] (ellipsis-species) to [out=150,in=-70] (species);
        \draw[draw=light-purple, arrow, dashed] (archaea) to [out=170,in=75] (species);
        \draw[draw=light-purple, arrow, dashed] (ellipsis-archaea) to [out=190,in=50] (archaea);
        
        \draw[draw=light-purple, arrow, dashed] (bacteria) to [out=-170,in=70] (species);        \draw[draw=light-purple, arrow, dashed] (ellipsis-bacteria) to [out=190,in=50] (bacteria);
        \draw[draw=light-purple, arrow, dashed] (eukaryote) to [out=180,in=-60] (species);
        \draw[draw=light-purple, arrow, dashed] (animal) to [out=180,in=60] (eukaryote);
        \draw[draw=light-purple, arrow, dashed] (fungus) to [out=180,in=-60] (eukaryote);
        \draw[draw=light-purple, arrow, dashed] (ellipsis-fungus) to [out=180,in=-40] (fungus);
        \draw[draw=light-purple, arrow, dashed] (plant) to [out=180,in=-70] (eukaryote);   
        \draw[draw=light-purple, arrow, dashed] (ellipsis-plant) to [out=180,in=-40] (plant);
        \draw[draw=light-purple, arrow, dashed] (ellipsis-eukaryote) to [out=160,in=-90] (eukaryote);
        \draw[draw=light-purple, arrow, dashed] (bird) to [out=180,in=60] (animal);  
        \draw[draw=light-purple, arrow, dashed] (ellipsis-bird) to [out=180,in=40] (bird);
        \draw[draw=light-purple, arrow, dashed] (mamma) to [out=180,in=-60] (animal);        
        \draw[draw=light-purple, arrow, dashed] (ellipsis-animal) to [out=190,in=70] (animal);
        \draw[draw=light-purple, arrow, dashed] (cat) to [out=180,in=60] (mamma);   
        \draw[draw=light-purple, arrow, dashed] (ellipsis-cat) to [out=180,in=40] (cat);
        \draw[draw=light-purple, arrow, dashed] (dog) to [out=180,in=-60] (mamma); 
        \draw[draw=light-purple, arrow, dashed] (ellipsis-dog) to [out=170,in=-70] (dog);
        \draw[draw=light-purple, arrow, dashed] (ellipsis-mamma) to [out=170,in=-70] (mamma);
        \draw[draw=light-purple, arrow, dashed] (boxer) to [out=180,in=60] (dog);   
        \draw[draw=light-purple, arrow, dashed] (saluki) to [out=180,in=-60] (dog);   
        
        \draw[draw=light-purple, arrow, dashed] (tool) to [out=130,in=-85] (thing);
        \draw[draw=light-purple, arrow, dashed] (ellipsis-tool) to [out=150,in=-70] (tool);
        \draw[draw=light-purple, arrow, dashed] (hair_slide) to [out=180,in=60] (tool);
        \draw[draw=light-purple, arrow, dashed] (ellipsis-hair_slide) to [out=180,in=40] (hair_slide);
        \draw[draw=light-purple, arrow, dashed] (device) to [out=180,in=-60] (tool);
        \draw[draw=light-purple, arrow, dashed] (camera) to [out=180,in=-60] (device);
        \draw[draw=light-purple, arrow, dashed] (ellipsis-camera) to [out=190,in=70] (camera);
        \draw[draw=light-purple, arrow, dashed] (ellipsis-device) to [out=190,in=50] (device);
        
        \draw[draw=light-purple, arrow, dashed] (instrument) to [out=180,in=60] (camera);
        \draw[draw=light-purple, arrow, dashed] (organ) to [out=180,in=-60] (camera);

        \draw[draw=light-purple, arrow, dashed] (vehicle) to [out=120,in=-95] (thing);
        \draw[draw=light-purple, arrow, dashed] (ellipsis-vehicle) to [out=150,in=-70] (vehicle);
        \draw[draw=light-purple, arrow, dashed] (airplane) to [out=180,in=60] (vehicle);
        \draw[draw=light-purple, arrow, dashed] (ellipsis-airplane) to [out=180,in=40] (airplane);
        
        \draw[draw=light-purple, arrow, dashed] (ship) to [out=180,in=-60] (vehicle);
        \draw[draw=light-purple, arrow, dashed] (aircraft_carrier) to [out=180,in=60] (ship);
        \draw[draw=light-purple, arrow, dashed] (ellipsis-ship) to [out=170,in=-75] (ship);
        
        \draw[draw=light-purple, arrow, dashed] (yawl) to [out=180,in=-60] (ship);

    \end{tikzpicture}
    \end{adjustbox}
    \caption{An abstract of the hierarchical relations existing in the ImageNet domain.}
    \label{fig:hierarchical relations imagenet}
\end{figure}